\documentclass[conference]{IEEEtran}
\IEEEoverridecommandlockouts
\usepackage{framed,multirow}
\usepackage{amssymb}
\usepackage{latexsym}
\usepackage{amsmath}
\usepackage{amsthm}
\usepackage{subcaption}
\usepackage{tabularx,booktabs}
\usepackage{tikz}
\usepackage{hyperref}

\newtheoremstyle{mytheorem}
  {5pt}
  {3pt}
  {\itshape}
  {}
  {\itshape\bfseries}
  {.}
  {.5em}
  {\thmname{#1}\thmnumber{{ }#2}%
   \thmnote{ {\the\thm@notefont(#3)}}}
\makeatother
\theoremstyle{mytheorem}

\newtheorem{thm}{Theorem}[section]

\newtheorem{defn}[thm]{Definition}

\usepackage{url}
\usepackage{xcolor}
\definecolor{newcolor}{rgb}{.8,.349,.1}
\definecolor{royalazure}{rgb}{0, 0, 0}

\def\BibTeX{{\rm B\kern-.05em{\sc i\kern-.025em b}\kern-.08em
    T\kern-.1667em\lower.7ex\hbox{E}\kern-.125emX}}
    
\makeatletter
\newcommand*\titleheader[1]{\gdef\@titleheader{#1}}
\AtBeginDocument{%
  \let\st@red@title\@title
  \def\@title{%
    \bgroup\normalfont\large\centering\@titleheader\par\egroup
    \vskip1.5em\st@red@title}
}

\title{Perturbation Analysis of Gradient-based \\Adversarial Attacks}
\titleheader{}

\author{
\IEEEauthorblockN{Utku Ozbulak$^{1,\,4}$ \quad Manvel Gasparyan$^{3,\,4}$ \quad Wesley De Neve$^{1,\,4}$  \quad Arnout Van Messem$^{2,\,4}$}
\IEEEauthorblockA{\small\textit{$^1$Department of Electronics and Information Systems, Ghent University, Belgium} \\
\textit{$^2$Department of Applied Mathematics, Computer Science and Statistics, Ghent University, Belgium} \\
\textit{$^3$Department of Data Analysis and Mathematical Modelling, Ghent University, Belgium}\\
\textit{$^4$Center for Biotech Data Science, Ghent University Global Campus, Republic of Korea}\\
\{utku.ozbulak,manvel.gasparyan,arnout.vanmessem,wesley.deneve\}@ugent.be}
\thanks{\textit{Preprint. Accepted for publication on Pattern Recognition Letters, 2020.}}
\thanks{DOI: \href{https://doi.org/10.1016/j.patrec.2020.04.034}{https://doi.org/10.1016/j.patrec.2020.04.034}.}
}

\begin{document}
\maketitle

\begin{abstract}
After the discovery of adversarial examples and their adverse effects on deep learning models, many studies focused on finding more diverse methods to generate these carefully crafted samples. Although empirical results on the effectiveness of adversarial example generation methods against defense mechanisms are discussed in detail in the literature, an in-depth study of the theoretical properties and the perturbation effectiveness of these adversarial attacks has largely been lacking. In this paper, we investigate the objective functions of three popular methods for adversarial example generation: the L-BFGS attack, the Iterative Fast Gradient Sign attack, and Carlini \& Wagner's attack. Specifically, we perform a comparative and formal analysis of the loss functions underlying the aforementioned attacks while laying out large-scale experimental results on ImageNet dataset. This analysis exposes (1) the faster optimization speed as well as the constrained optimization space of the cross-entropy loss, (2) the detrimental effects of using the signature of the cross-entropy loss on optimization precision as well as optimization space, and (3) the slow optimization speed of the logit loss in the context of adversariality. Our experiments reveal that the Iterative Fast Gradient Sign attack, which is thought to be \textit{fast} for generating adversarial examples, is the worst attack in terms of the number of iterations required to create adversarial examples in the setting of equal perturbation. Moreover, our experiments show that the underlying loss function of Carlini \& Wagner's attack, which is criticized for being substantially slower than other adversarial attacks, is not that much slower than other loss functions. {\color{royalazure}Finally, we analyze how well neural networks can identify adversarial perturbations generated by the attacks under consideration, hereby revisiting the idea of adversarial retraining on ImageNet.}
\end{abstract}

\section{Introduction}
\label{Introduction}
With the groundbreaking results obtained by deep learning models for different types of machine learning problems \cite{DBLP:journals/corr/RedmonDGF15, resnet}, studies that investigate the weaknesses of these models also gained traction \cite{Goodfellow-expharnessing, on_detecting, moosavi2016deepfool, DBLP:journals/corr/PapernotMG16}. From these studies, it became clear that deep learning models are vulnerable to gradient-based attacks that create so-called (negative) adversarial examples. To counter these adversarial attacks, a number of defense techniques have been proposed, only to be falsified by follow-up studies \cite{DBLP:journals/corr/CarliniW17}.

As more attacks are proposed with each study, hereby showing that the newly introduced methods are \textit{superior} to their predecessors by mostly empirical means, a mathematical explanation as to why the proposed attacks are \textit{better} than others remains lacking. A similar observation can be made regarding the availability of rigorous comparative analyses of the proposed attacks. In this study, we aim at formally explaining a number of properties of commonly used gradient-based attacks, namely, the L-BFGS attack (L-BFGS), Iterative Fast Gradient Sign (IFGS), and Carlini \& Wagner's attack (CW). We investigate (1) why some of these attacks are faster than others when generating adversarial examples~\cite{IFGS} and (2) why some of these attacks are creating \textit{stronger} adversarial examples that are more robust against state-of-the-art defense mechanisms~\cite{CW_Attack}. While doing so, we also analyze the effectiveness of the perturbation generated by different loss functions, in a setting where the total amount of perturbation is held constant, so to allow for a fair comparison. To the best of our knowledge, this is the first study that makes use of the aforementioned setting in order to analyze the effectiveness of perturbations that have been generated by adversarial attacks. In this context, we first lay out the theoretical advantages and disadvantages of the aforementioned attacks, subsequently providing empirical results that are in support of our theoretical findings, and where the empirical results have been obtained by performing large-scale experiments in the image domain using the ImageNet dataset~\cite{ILSVRC15:rus}. {\color{royalazure}Finally, we analyze whether or not various deep neural network architectures can distinguish adversarial examples from genuine images.}

Our paper is organized as follows. In Section~\ref{Notation and Framework}, we briefly outline our mathematical notations. In Section~\ref{Adversarial Example Generation Methods}, we first provide a detailed review of the different methods used in this study for adversarial example generation, subsequently generalizing their objective functions in terms of how the perturbation is generated. In Section~\ref{Loss Function Properties}, we analyze the mathematical properties of the objective functions studied as. In Section~\ref{Experiments}, we present experimental results. Finally, we provide conclusions and directions for future research in Section~\ref{Conclusion}.

\section{Mathematical Notations}
\label{Notation and Framework}

\begin{itemize}
\item $\mathbf{X}$ \textemdash \, an arbitrary image represented as a 3-D tensor (depth $\times$ width $\times$ height), with {\color{royalazure}values} in the range of $[0, 1]$.

\item $\mathbf{y} = g(\theta, \mathbf{X})$ \textemdash \, a classification function that links $\mathbf{X}$ to an output vector $\mathbf{y}$ of size $M$, containing predictions made by a neural network that comes with parameters $\theta$ and that does not contain a final softmax layer. $M$ denotes the total number of classes used. The $k$-th element of this vector is referred to as $g(\theta, \mathbf{X})_k$.

\item $S(g(\theta, \mathbf{X})) = \left [ \dfrac{e^{g(\theta, \mathbf{X})_t}}{\sum_{m=1}^{M}e^{g(\theta, \mathbf{X})_m}} \right ]_{t=1}^{M} \,$ \textemdash \, the vector generated by the softmax function. This function converts the logits produced by the neural network into probabilistic form.

\item $J(g(\theta, \mathbf{X})_{c}) =  - \log \left( \dfrac{e^{g( \theta, \mathbf{X})_{c} }}{ \sum_{m=1}^{M} e^{g( \theta , \mathbf{X})_{m}} } \right)$ \textemdash \, the cross-entropy function (CE), which calculates the negative logarithmic loss of the softmax prediction made by a neural network for a class $c$.

\item $\nabla_{x} g(\theta, \mathbf{X})$ \textemdash \, the partial derivative of a neural network $g$ with respect to an input $\mathbf{X}$. 
\end{itemize}

\section{Adversarial Example Generation Methods}
\label{Adversarial Example Generation Methods}
In the field of adversarial examples, a variety of attacks exists \cite{CW_Attack,Goodfellow-expharnessing,IFGS,PGD_attack,moosavi2016deepfool,JSMA,LBFGS}. However, in this study, we only consider targeted adversarial attacks, given that this type of attacks is typically preferred over untargeted attacks when assessing the robustness of defense methods \cite{DBLP:journals/corr/CarliniW17}. In that regard, we focus on three popular methods that are frequently used for generating adversarial examples: the \textit{L-BFGS attack}, the \textit{Iterative Fast Gradient Sign attack}, and  \textit{Carlini \& Wagner's attack}. 

\textbf{L-BFGS}: Box-constrained L-BFGS, as introduced by \cite{LBFGS}, is one of the first widely used adversarial attacks. This attack aims at finding an adversarial example $\mathbf{X}'$ that is similar to the input $\mathbf{X}$ under the $L_{2}$ distance, but that gets assigned a different label by the model. \cite{LBFGS} expressed this problem in a $d$-dimensional setting as follows: 
\begin{equation}
\begin{aligned}
& \underset{}{\text{minimize}}
& & b \cdot || \mathbf{X} - \mathbf{X}' ||^{2}_{2} - \ell(\mathbf{X}') \,, \\
& \text{such that}
& & {\color{royalazure}\mathbf{X}'}\in [0, 1]^{d} \,.
\end{aligned}
\label{eq:lbfgs}
\end{equation}
This minimization problem is repeatedly solved for multiple values of $b$, using bisection search to find optimal perturbations. A common choice for the loss function $\ell$, and one we will also follow, is the cross-entropy function. 

\textbf{Iterative Fast Gradient Sign}: IFGS \cite{IFGS}, also referred to as the \textit{basic iterative method}, finds its origin in the \textit{Fast Gradient Sign} (FGS) method proposed by \cite{Goodfellow-expharnessing}. The main objective of both FGS and IFGS is to generate adversarial examples quickly. As such, IFGS updates an image using the sign of the gradient at each iteration as: 
\begin{align} \label{eq:ifgs}
\mathbf{X}_{n+1} =  Clip_{\mathbf{X}, \epsilon}(  \mathbf{X}_{n} - \alpha \ \text{sign} (\nabla_x J(g(\theta,\mathbf{X}_n)_c)  \,.
\end{align}
In the above equation, the perturbation is generated using the signature of the cross-entropy loss, which we will refer to as the CE-sign loss. The parameter $\epsilon$ controls the maximally allowed amount of perturbation per pixel, $\alpha$ determines the amount of CE-generated perturbation added by each iteration, and the clipping function ensures that the resulting adversarial example remains a valid image. In this study, we opted for IFGS over FGS, given that IFGS has been shown to produce stronger adversarial examples \cite{IFGS}. CE-sign loss is also used by Projected Gradient Descent Attack later proposed by~\cite{PGD_attack}.


\textbf{Carlini \& Wagner}: \cite{CW_Attack} proposed a heavily optimized attack which produces strong adversarial examples that can bypass defense mechanisms easily.
The $L_2$ version of this attack, which was also used by~\cite{DBLP:journals/corr/CarliniW17} to test the robustness of defense mechanisms, is defined as follows:
\begin{align}
\text{miminize} \quad & ||\mathbf{X} - (\mathbf{X} + \delta)||_{2}^{2} +  \; \ell(\mathbf{X} + \delta)\,,
\end{align}
where this attack attempts to find a small perturbation $\delta$ that is sufficient to change the prediction made by the model when it is added to the input, while keeping the $L_2$ distance between the original image $\mathbf{X}$ and the perturbed image $\mathbf{X}' = \mathbf{X} + \delta$ minimal.


We will use the loss they preferred in their later work to evaluate multiple defense mechanisms:
\begin{align} \label{eq:CW_2}
\ell (\mathbf{X}') = \max \big(  \max \{ g(\theta, \mathbf{X}')_i : i\neq c\} - g(\theta, \mathbf{X}')_c, -  \kappa \big) \,,
\end{align}
where $\max \{ g(\theta, \mathbf{X}')_i : i\neq c\} - g(\theta, \mathbf{X}')_c $ compares the logit value of target class $c$ with that of the next-most-likely class $i$. The constant $\kappa$ can be used to adjust the \textit{strength} of the produced adversarial example. 

\cite{CW_Attack} refer to the loss function shown in Equation (\ref{eq:CW_2}) as \textit{logit loss}, which is also commonly known as \textit{using the logits} to generate adversarial examples. However, throughout this paper, and for the sake of clarity, we will use (1) the term \textit{logit loss} when maximizing a single target class using logits and (2) the term \textit{multi-target logit loss} (abbreviated as M-logit loss) when we refer to the loss function of CW (the latter focuses on two classes, namely the target class and the next-most-likely class).

{\color{royalazure}\textbf{Generalizing Objective Functions}:} Although the objective functions presented in this section differ in multiple ways and are complex in nature, it is possible to generalize them in terms of how the perturbation is generated. During each step, these methods essentially aim at finding a perturbation that increases the prediction likelihood of the target class. In order to investigate properties of interest of the different methods for adversarial example generation, like robustness and speed, we first rewrite the objective functions of L-BFGS, IFGS, and CW in a similar format, with this format explicitly displaying the source of the generated perturbation. As mentioned before, to generate the perturbation, L-BFGS uses CE loss, IFGS uses CE-sign loss, and CW uses M-logit loss. In order to display the source of the perturbation, we rewrite the aforementioned attacks as follows:

\begin{flalign}
&\mathbf{X}_{n+1} = \,\, \zeta_1 \Big(  \mathbf{X}_{n}  +  \gamma_1 \, \big( \nabla_{x} J(g(\theta, \mathbf{X}_{n})_{c}) \big) \Big) \,, \label{eq:revised_lbfgs}\\
&\mathbf{X}_{n+1} = \,\, \zeta_2 \Big(  \mathbf{X}_{n}  +  \gamma_2 \big( \, \text{sign} \, \big( \nabla_{x} J(g(\theta, \mathbf{X}_{n})_{c}) \big) \big) \Big)  \,, \label{eq:revised_ifgs}\\
&\mathbf{X}_{n+1} = \,\, \zeta_3 \Big( \mathbf{X}_{n}  + \gamma_3 \ (\nabla_x g(\theta,\mathbf{X}_{n})_{c}) +  \gamma_4 \ ( \nabla_x g(\theta,\mathbf{X}_{n})_{c*} \big) \Big)\,,
\label{eq:revised_cw}
\end{flalign}
where Equations~(\ref{eq:revised_lbfgs}),~(\ref{eq:revised_ifgs}), and~(\ref{eq:revised_cw}) correspond to {\it adversarial optimizations} performed by L-BFGS, IFGS, and CW, respectively. In the above equations, $c$ refers to the target class, and the $\gamma$-function, which is unique for each method, is used to satisfy the properties of the generated perturbation and thus encompasses, e.g., the perturbation multiplier and the clipping function. The $\zeta$-function, in its turn, is used to ensure that the generated adversarial examples satisfy various constraints, such as box-constraints and $L_2$ distance minimization. In Equation~(\ref{eq:revised_cw}), the second-most-likely target class is referred to as $c^{*}$.

By writing the adversarial optimizations in this way, it becomes clear that, in order to analyze the properties of adversarial example generation methods, the properties of the loss functions must be investigated. Therefore, in the next section, we will analyze the differences between the cross-entropy loss $\nabla_{x} J(g(\theta, \mathbf{X})_{c})$ as used by L-BFGS, the cross-entropy signature loss $sign(\nabla_{x} J(g(\theta, \mathbf{X})_{c}))$ as used by IFGS, and the logit loss $\nabla_{x} g(\theta, \mathbf{X})_{c}$, whose extension M-logit loss is used by CW.

\begin{figure}[t]
\centering
\includegraphics[scale=.4]{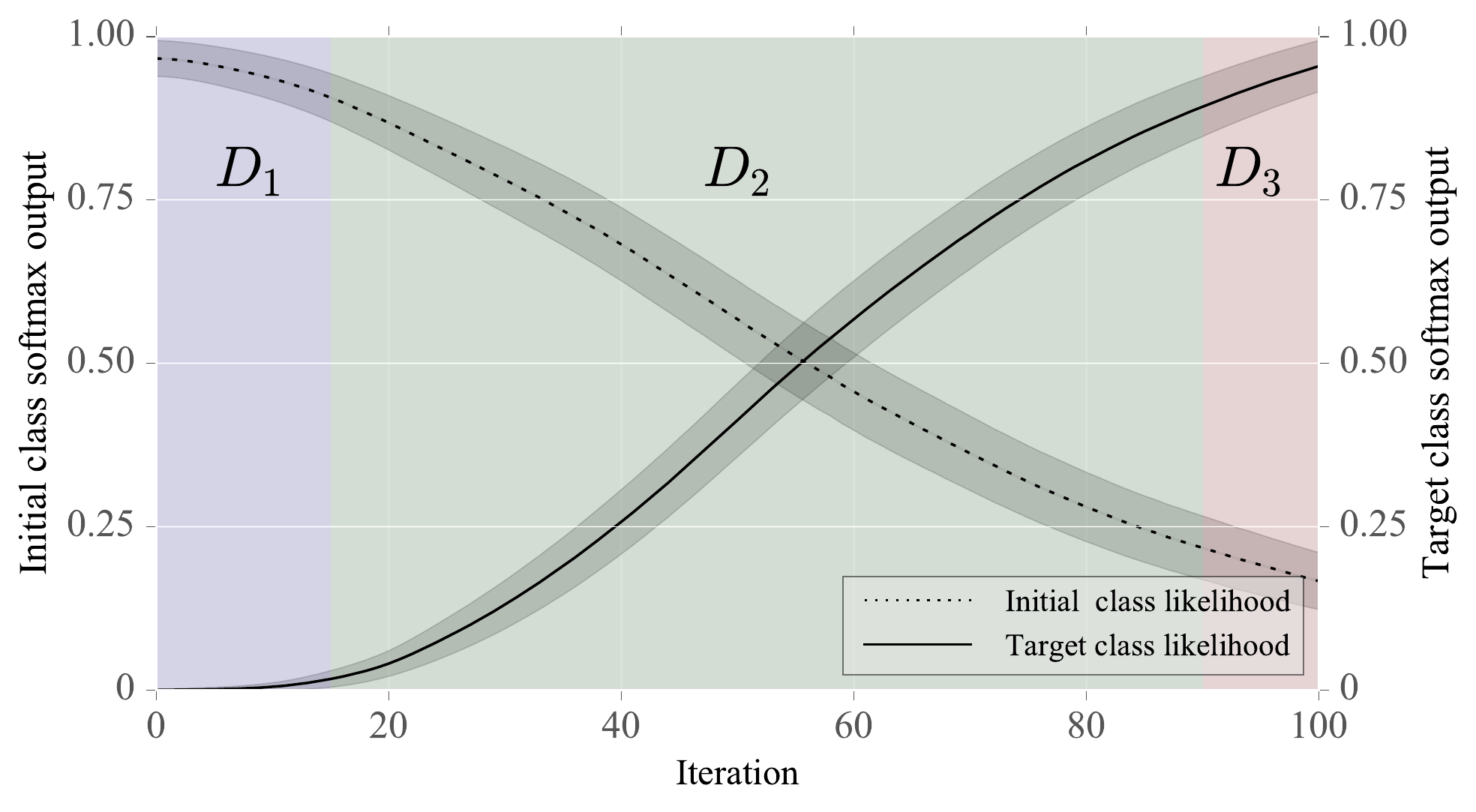}
\caption{The prediction likelihood of both the initial class and the target class is displayed throughout $100$ iterations of adversarial optimization. The lines represent the mean and the shaded areas represent the $95\%$ confidence interval of likelihood values that have been obtained for $1,000$ adversarial examples. The subspaces defined in Definition~\ref{def:subspace_definition} are highlighted as blue, green, and red areas, representing $D_1$, $D_2$, and $D_3$, respectively.}
\label{fig:subspace_graph}

\end{figure}

\section{Properties of Loss Functions in Adversarial Settings}
\label{Loss Function Properties}

In the previous section, we wrote the objective functions of different methods for adversarial example generation in such a way that the techniques for perturbation generation are comparable. In order to understand the limitations and advantages of these techniques, we analyze their behavior for various outputs of $g(\theta, \mathbf{X})$, hereby investigating the gradient produced.

A data point selected for adversarial optimization is assumed to be correctly classified first, and the aim is to eventually change the prediction, so to force the machine learning model under consideration to incorrectly classify the data point selected (often with high confidence). Consider Fig.~\ref{fig:subspace_graph}, which shows the mean prediction likelihood for both the initial class and the target class. This mean prediction likelihood was obtained by creating $1,000$ adversarial examples, applying IFGS to samples taken from the ImageNet dataset. A pretrained ResNet-50~\cite{resnet} in white-box settings was used for this small experiment. 

As can be seen from Fig.~\ref{fig:subspace_graph}, the adversarial optimization causes the prediction likelihood of the initial class to decrease and the prediction likelihood of the target class to increase. Note that the method used for adversarial optimization does not change the overall outcome, but only the rate of change in the likelihoods. Indeed, the ImageNet samples are, in the end, mostly classified as the target class, with confidence values on average higher than $0.9$.

Based on observations made for the above experiment, we adopt the following definition.

\begin{defn}
\label{def:subspace_definition}
Let $\mathbf{y} = g(\theta, \mathbf{X})$ be a neural network that maps an input $\mathbf{X}$ to an output vector $\mathbf{y}$ and let $c$ be the target class. Based on the softmax output $S(g(\theta, \mathbf{X}))$ of the classification result, we define three subspaces $D_{\{1,2,3\}}$:

\begin{itemize}
    \item $\mathbf{D_1}$\,\textemdash \,\,The input is classified with high confidence as some class $r$ other than the target class $c$:    $\forall \mathbf{X}_1 \in D_1$, $\max (S(g(\theta, \mathbf{X}_1))) \geq 0.9$ and $\arg\max g(\theta, \mathbf{X}_1)  = r,  r \neq c$.
    \item $\mathbf{D_2}$\,\textemdash \,\,The input is classified with relatively low confidence for any class: $ \forall \mathbf{X}_2 \in D_2$, $\max (S(g(\theta, \mathbf{X}_2))) < 0.9$.
    \item $\mathbf{D}_3$\,\textemdash \,\,The input is classified with high confidence as the target class $c$: $ \forall \mathbf{X}_3 \in D_3$, $S(g(\theta, \mathbf{X}_3)) \geq 0.9$ and $\arg\max S(g(\theta, \mathbf{X}_3))  = c$.
\end{itemize}
\end{defn}

The subspaces listed in Definition~\ref{def:subspace_definition} are visualized in Fig.~\ref{fig:subspace_graph}. A data point that is selected for adversarial optimization is assumed to be correctly classified (more-often-than-not with high confidence), and hence resides in $D_1$. The aim of the adversarial example generation methods we discussed in Section~\ref{Adversarial Example Generation Methods} is to move this data point from $D_1$ to $D_3$ in an iterative manner, where it is eventually classified as the target class. During this optimization process, the data point under consideration inevitably passes through $D_2$.

In what follows, we analyze the objective functions in more detail, with the goal of answering the following question: \textit{``what makes an objective function beneficial or detrimental when generating adversarial examples?''} In Section~\ref{Adversarial Example Generation Methods}, we showed that it is possible to distinguish between adversarial example generation methods by their usage of CE, CE-sign, and logits. As a result, we investigate the limitations and advantages of CE loss and CE-sign loss against logit loss in the context of adversarial example generation.

\subsection{Cross-Entropy Loss}
The cross-entropy loss was, thanks to its desirable properties, one of the first loss functions used to generate adversarial examples~\cite{LBFGS}. The following theorem describes the behavior of the CE loss for the subspaces $D_{\{1,2,3\}}$.

\begin{thm} \label{thm:1}
Let $\mathbf{y} =g(\theta, \mathbf{X})$ be a neural network, represented by a differentiable function $g$ that maps an input $\mathbf{X}$ to an output vector $\mathbf{y}$, and assume that adversarial examples are generated using the cross-entropy function $J(g(\theta, \mathbf{X})_{c})$.

Based on the subpaces described in Definition~\ref{def:subspace_definition}, the gradient $\nabla_x J(g(\theta, \mathbf{X})_{c})$ of the cross-entropy function can be approximated as follows:

\begin{itemize}
\item $\forall \mathbf{X}_1 \in D_1,$
\begin{flalign} 
&\lim_{\mathbf{X} \to \mathbf{X}_{1}} \nabla_{x} J(g(\theta, \mathbf{X})_{c}) = \nabla_{x} g(\theta , \mathbf{X}_1)_{r} - \nabla_{x} g(\theta , \mathbf{X}_1)_{c} \,.&
\end{flalign}

\item $\forall \mathbf{X}_2 \in D_2,$
\begin{flalign}
\label{eq:d2}
&\lim_{\mathbf{X} \to \mathbf{X}_{2}} \nabla_{x} J(g(\theta, \mathbf{X})_{c}) = \\
&\sum_{m \in I\setminus \{c\}} \beta_m  \Big(\nabla_{x} g(\theta , \mathbf{X}_2)_{m} 
 - \nabla_{x} g(\theta , \mathbf{X}_2)_{c}\Big)\,,\nonumber 
\end{flalign}
with constants $\beta_1,\ldots,\beta_M$ subject to \phantom{asdasd} $\sum_{m \in I\setminus \{c\}} | \beta_m | < 1$, where $I = \{1, 2, \ldots, M\}$.
\newline

\item $\forall \mathbf{X}_3 \in D_3,$
\begin{flalign} 
\lim_{\mathbf{X} \to \mathbf{X}_{3}} \nabla_{x} J(g(\theta, \mathbf{X})_{c}) =  0\,.&
\end{flalign}

\end{itemize}
\end{thm}

Theorem~\ref{thm:1} is proved in the supplementary material.

Let us now discuss the outcome of Theorem~\ref{thm:1} for the data points that lie in $D_1$, $D_2$, and $D_3$, respectively.

\textbf{Optimization for Data Points in $D_1$}\,\textemdash\,In this subspace, where the data point under consideration is classified with high confidence as any class other than the targeted one, using the gradient of the CE loss produced by gradient descent, the likelihood of the current class $r$ will be minimized and, at the same time, the likelihood of the target class $c$ will be maximized. This outcome naturally increases the optimization speed compared to only maximizing the target class likelihood, given the fact that the data points in $D_1$ are classified as $r$.

CW, even though regarded as the attack that produces the strongest adversarial examples, is also criticized for being substantially slower (i.e., computationally more expensive) than IFGS and L-BFGS \cite{Goodfellow:2018:MML:3234519.3134599}. This is not only because the algorithm itself is complex (i.e., incorporating mechanisms like multiple-starting-point gradient descent), but also because CW uses M-logit loss instead of CE loss. As we have stated above, CE loss, compared to logit loss, has a faster speed at the start of the optimization (i.e., for the data points in $D_1$) and is thus able to change the prediction of the data point at hand faster.

We will show empirical results on the ImageNet dataset in support of the aforementioned claims in Section~\ref{Experiments}.

\textbf{Optimization for Data Points in $D_2$}\,\textemdash\,A data point located in subspace $D_2$ is classified with low confidence as one of the available classes. Once here, the optimization speed starts to decrease because $\sum_{m \in I\setminus \{c\}} | \beta_m | < 1$, as can be seen from Equation~(\ref{eq:d2}). This is different from the optimization in $D_1$, where, provided that the equation is written in the same format, the gradient multiplier is $1$.

\textbf{Optimization for Data Points in $D_3$}\,\textemdash\,In subspace $D_3$, the data point under consideration is already classified as the target class with high confidence. In this case, the loss generated with the cross-entropy approaches zero, which means that the data point under consideration cannot be further optimized. We will refer to this mathematical constraint of the CE loss in an adversarial setting for data points that lie in $D_3$ as the mathematical limit of the optimization space of the CE loss.

The biggest difference between logit loss-based optimization methods (e.g., CW) and CE loss-based optimization methods (e.g., FGS, IFGS, and L-BFGS) can be observed for the points that lie in $D_3$. As we have shown in Theorem~\ref{thm:1}, CE loss cannot generate any further gradient signal as soon as the data point under consideration is classified with high confidence as the target class. This constraint, as will become clear from the examples, does not apply to logit loss-based optimization methods, explaining why CW is able to create arguably \textit{stronger} adversarial examples.

\subsection{Cross-Entropy Sign Loss}
We also experimented with taking the signature of the gradient of the CE loss, a method that was shown to be successful in generating adversarial examples~\cite{Goodfellow-expharnessing, IFGS}. Although it is not clear whether or not this method was used in the aforementioned papers to overcome the subspace limitation of the CE loss that we previously laid out, we will show that taking the signature does not alleviate the subspace limitation of the CE loss in adversarial settings, and that it also brings along its own detrimental properties.

\textbf{Limited Optimization Space}\,\textemdash\,Since the limit of the gradient generated with the CE loss, for the data points in $D_3$, approaches zero, multiplying this gradient with a large number is not sufficient to move the data point under consideration deeper into $D_3$. In computational settings with limited decimal precision, when, in absolute value, the largest element of the gradient matrix $\nabla_{x} J(g(\theta, \mathbf{X})_{c})$ for data points that reside in $D_3$ becomes smaller than the numerical precision used by adversarial optimization (for example, $1e-8$ for single precision or $1e-16$ for double precision), then $\nabla_{x} J(g(\theta, \mathbf{X})_{c}) \equiv 0$. As a result, taking the signature of the gradient generated with CE loss, and using it instead of the gradient itself, will not further optimize the data points that lie in $D_3$.

\textbf{Eliminated Rate-of-Change Among Elements}\,\textemdash\,Since both FGS and IFGS take the signature of the gradient, only the direction of the optimization is retained for individual elements of the gradient, and information about the rate of change is lost. As a result, the optimization step size for each element becomes identical, which can hinder the optimization speed. We will show in the upcoming section that, on average, FGS and IFGS perform worse than direct usage of gradients in terms of the total amount of perturbation added.

\section{Experiments}
\label{Experiments}
In support of our mathematical observations, we performed two sets of large-scale experiments on the ImageNet dataset~\cite{ILSVRC15:rus}.
{\color{royalazure}
The aim of the first set of experiments is to reveal the properties of perturbations, as generated by the attacks covered in Section~\ref{Adversarial Example Generation Methods}, and the aim of the second set of experiments is to discover the degree of detectability of these adversarial perturbations by neural networks.
}
\begin{figure*}[ht]
\centering
\begin{subfigure}{0.3\textwidth}
\includegraphics[width=5.6cm]{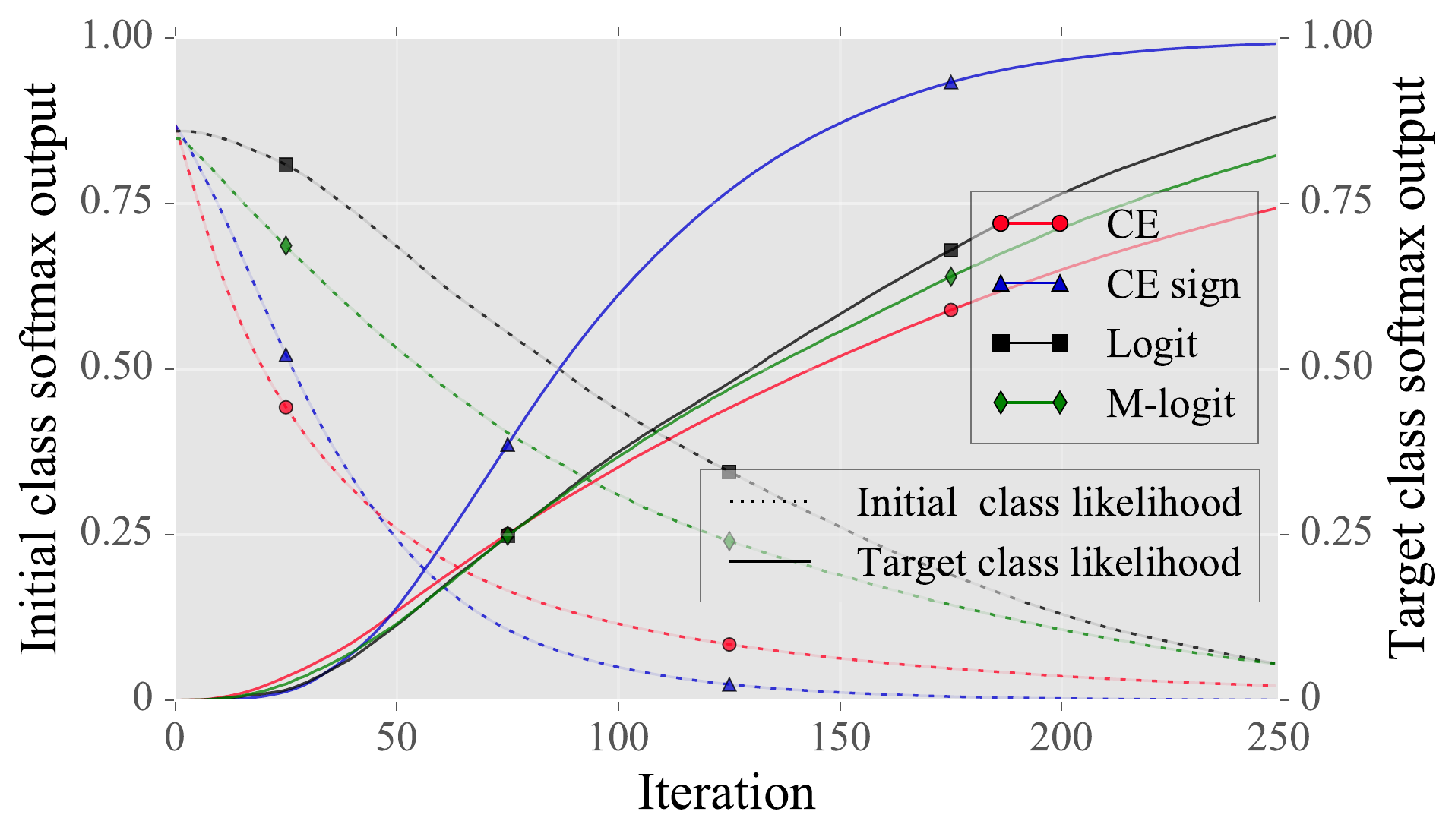}
\caption{Softmax output}
\label{fig:eqmul-line-1}
\end{subfigure}
\begin{subfigure}{0.3\textwidth}
\includegraphics[width=5.6cm]{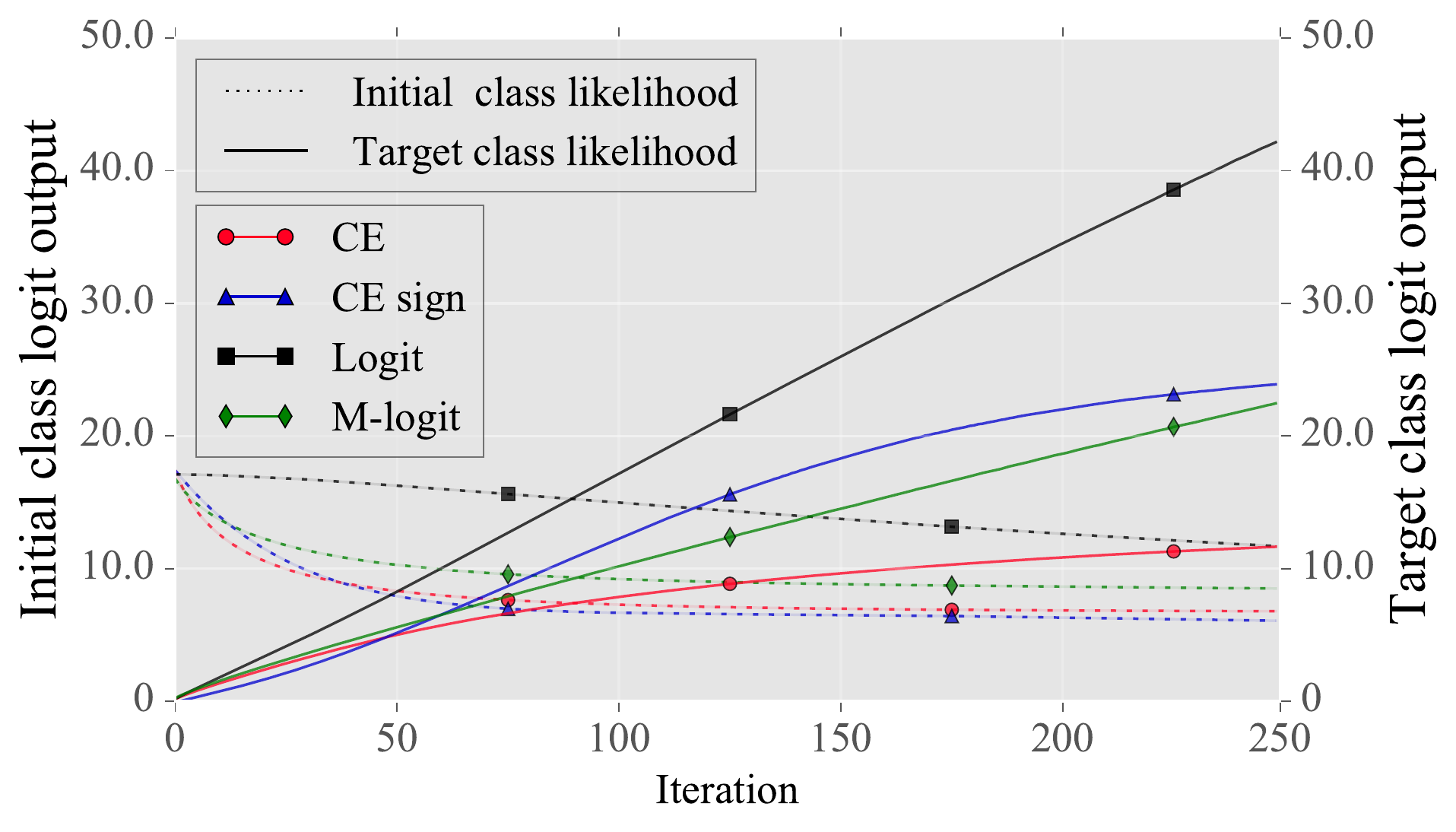}
\caption{Logit output}
\label{fig:eqmul-line-2}
\end{subfigure}
\begin{subfigure}{0.3\textwidth}
\includegraphics[width=5.6cm]{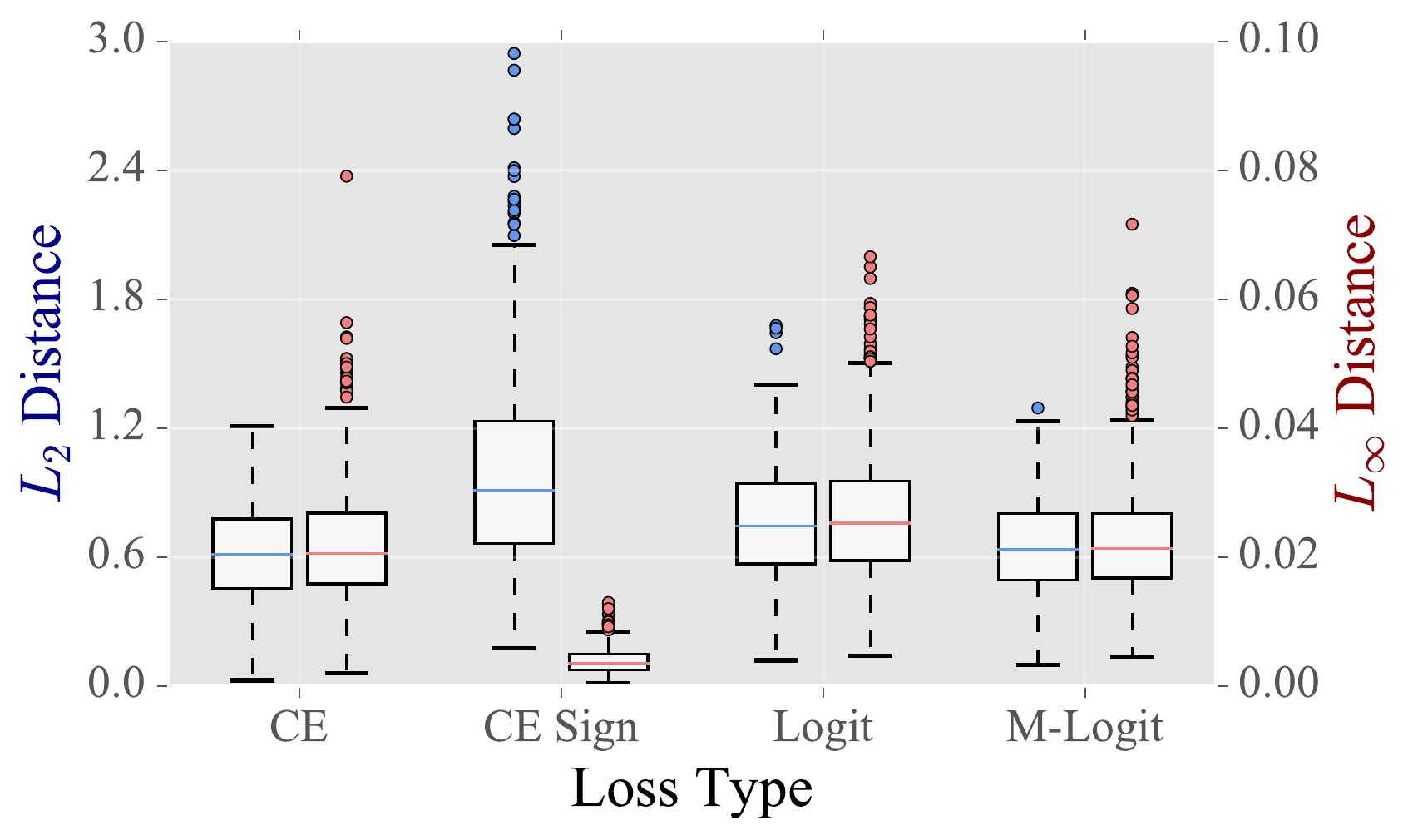}
\caption{Least amount of required perturbation}
\label{fig:eqmul-line-3}
\end{subfigure}

\caption{(a) The softmax output and (b) the logit prediction for both the initial class and the target class is given for adversarial optimization over $250$ iterations, for the experiment \textit{``equal multiplier''}. The values correspond to the mean value of $1,000$ samples, taken from ImageNet validation set. (c) The least amount of total added perturbation required to change the prediction of the data points under consideration (the same data points are presented in (a) and (b)). Best viewed in color.}
\label{fig:eqmul-opt}
\end{figure*}

\begin{figure*}[ht]
\centering
\begin{subfigure}{0.3\textwidth}
\includegraphics[width=5.6cm]{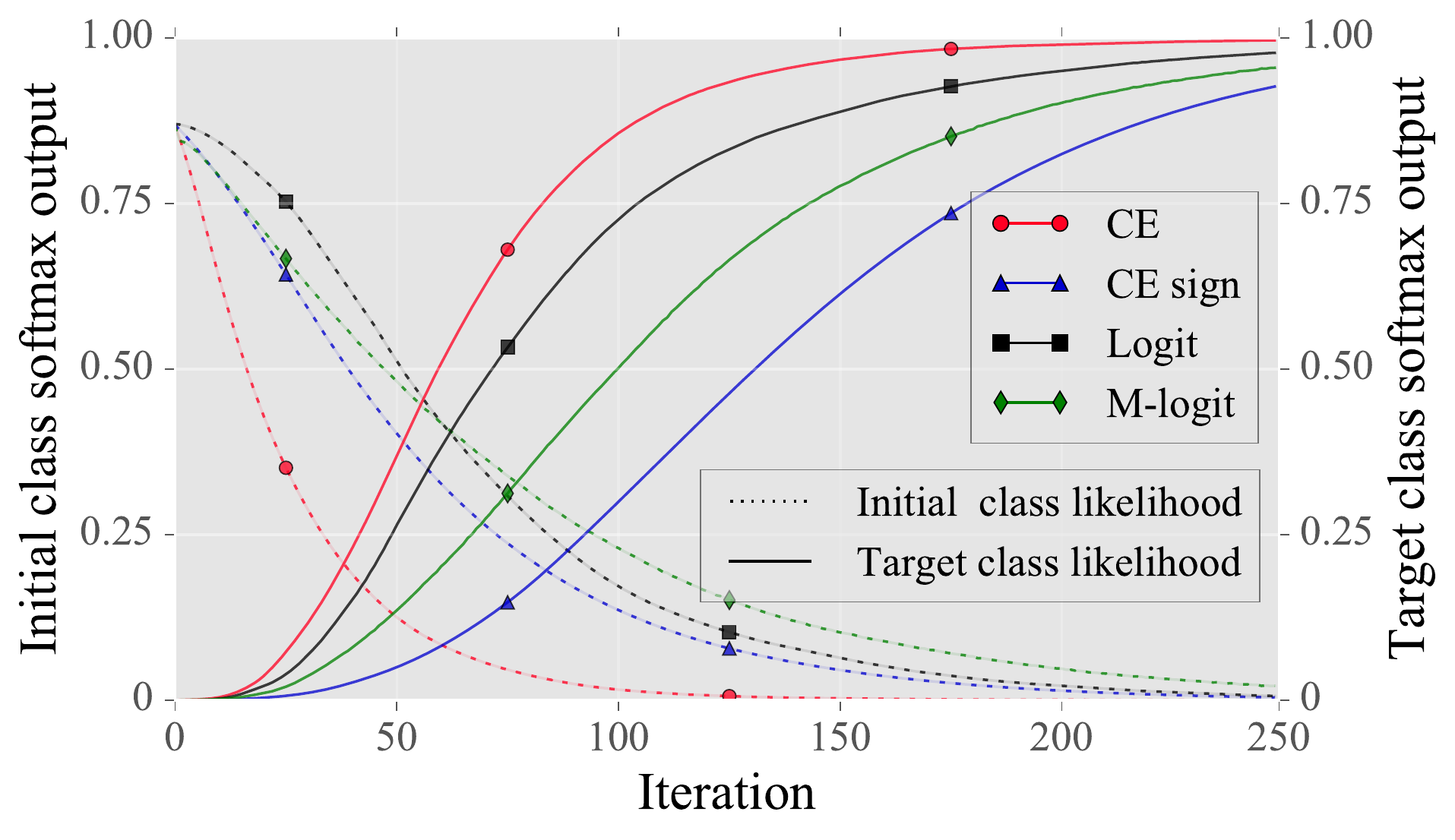}
\caption{Softmax output}
\label{fig:eqper-line-1}
\end{subfigure}
\begin{subfigure}{0.3\textwidth}
\includegraphics[width=5.6cm]{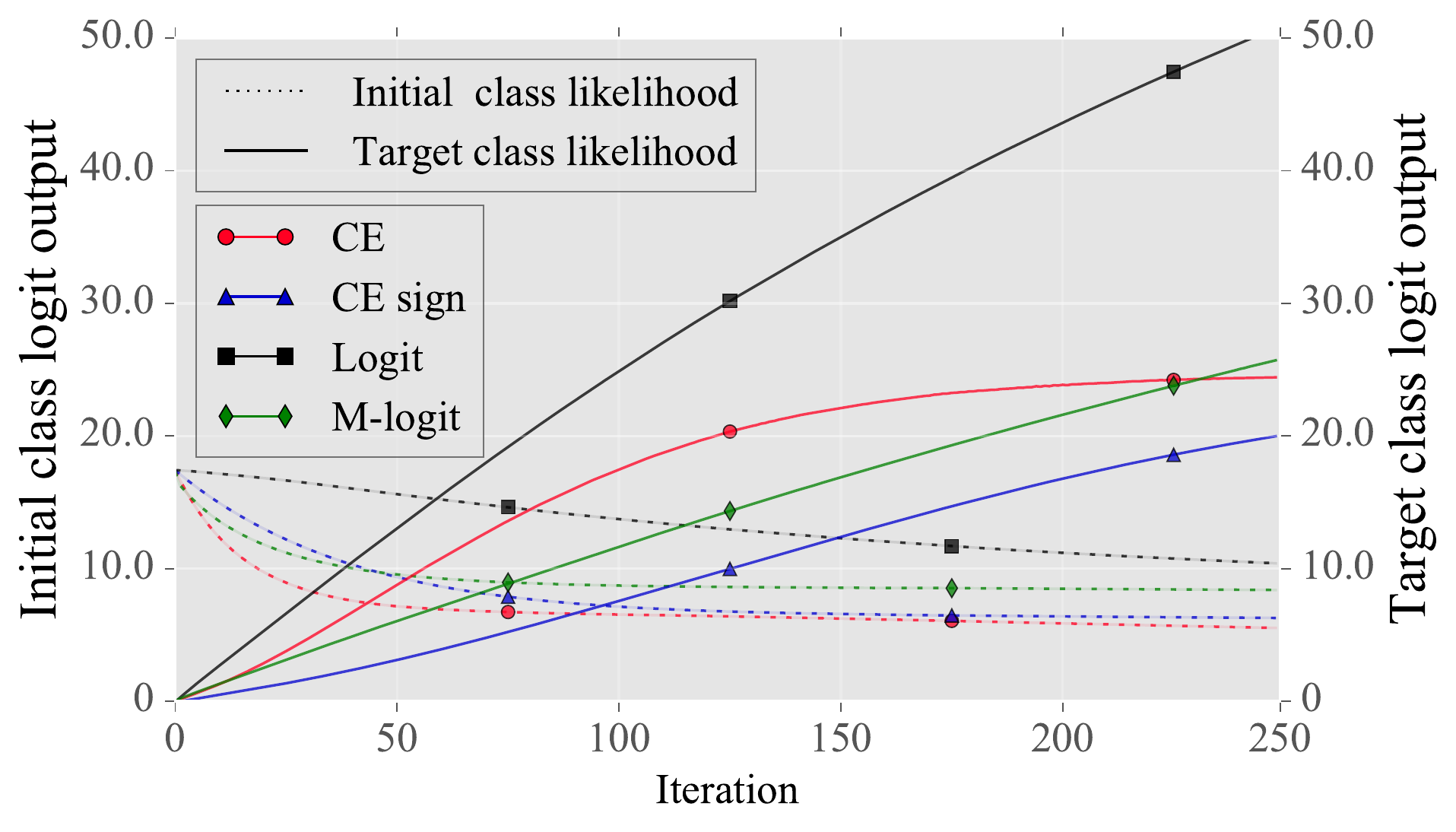}
\caption{Logit output}
\label{fig:eqper-line-2}
\end{subfigure}
\begin{subfigure}{0.3\textwidth}
\includegraphics[width=5.6cm]{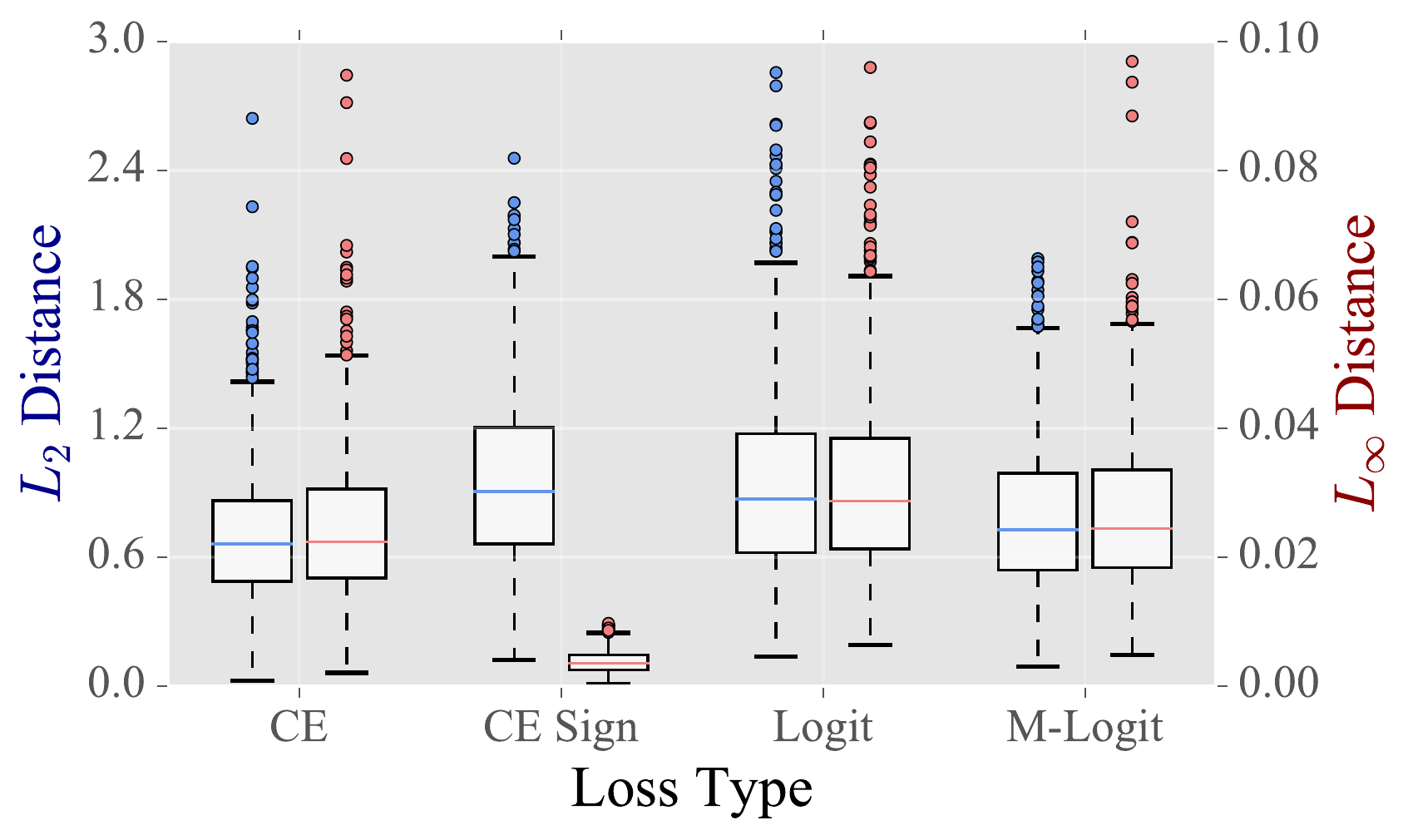}
\caption{Least amount of required perturbation}
\label{fig:eqper-line-3}
\end{subfigure}

\caption{(a) The softmax output and (b) the logit prediction for both the initial class and the target class is given for adversarial optimization over $250$ iterations, for the experiment \textit{``equal perturbation''}. The values correspond to the mean value of $1,000$ samples, taken from ImageNet validation set. (c) The least amount of total added perturbation required to change the prediction of the data points under consideration (the same data points are presented in (a) and (b)). Best viewed in color.}
\label{fig:eqper-opt}

\end{figure*}

\subsection{Experimental Setup}
\label{Experimental Setup}

We rely on the following method to generate the adversarial examples:
\begin{flalign}
\mathbf{X}_{n+1} & = \,\, \zeta \Big(  \mathbf{X}_{n}  +  \alpha \, \mathbf{P}_n \Big) \label{eq:imagenet_adv} \,,& \\
\textit{CE loss:} \,\,  \mathbf{P}_n &= \nabla_{x} J(g(\theta, \mathbf{X}_{n})_{c})  \,,& \label{eq:imagenet_ce} \\
\textit{CE-sign loss:} \,\,  \mathbf{P}_n &= \text{sign} \, \big( \nabla_{x} J(g(\theta, \mathbf{X}_{n})_{c}) \big) \,,& \label{eq:imagenet_ifgs} \\
\textit{Logit loss:} \,\, \mathbf{P}_n & = \nabla_x g(\theta,\mathbf{X}_{n})_{c} \label{eq:imagenet_logit} \,,& \\
\textit{M-logit loss:} \,\, \mathbf{P}_n & = \alpha' \,  \nabla_x g(\theta,\mathbf{X}_{n})_{c} +  \alpha^{*} \,  \nabla_x g(\theta,\mathbf{X}_{n})_{c*} \,, \label{eq:imagenet_mlogit}& \\
& \text{with}\,  g(\theta,\mathbf{X}_{n})_{c}- g(\theta,\mathbf{X}_{n})_{c*} = \kappa \,, \label{CW_const_2}
\end{flalign}
where Equation~(\ref{eq:imagenet_adv}) represents the general approach to create adversarial examples, with $\mathbf{P}_n$ the perturbation added during each iteration, and $\alpha$ the perturbation multiplier. In this specific setting, the $\zeta$-function is, again, used to ensure that the generated adversarial examples satisfy the box constraints. Equations~(\ref{eq:imagenet_ce}) to (\ref{eq:imagenet_mlogit}) describe the source of perturbation for the different methods. Note that we also include the logit loss in order to continue the comparative analysis between the logit loss, the CE loss, and CE-sign loss. In the case of the \textit{M-logit loss}, we follow the work of \cite{CW_Attack} and select $\kappa=20$. To ensure that Equation~(\ref{CW_const_2}) holds for $\kappa=20$, the multipliers $\alpha'$ and $\alpha^{*}$ are used to adjust the level of perturbation generated by their respective sources.

\subsection{Perturbation Analysis of Adversarial Attacks}
\label{Perturbation Analysis}
{\color{royalazure}
To investigate the properties of the perturbation generated by the different loss functions discussed in Section~\ref{Adversarial Example Generation Methods}, we conduct two experiments on ImageNet using different settings: (1) keeping the perturbation multiplier constant and (2) keeping the perturbation itself equal. We generate $1,000$ adversarial examples for each attack and for each experiment, using a pretrained ResNet-50 network~\cite{resnet} in white-box settings, hereby performing a detailed analysis of the properties of the generated perturbations.
}
These two large-scale experiments generalize a simulation, of which the details can be found in the supplementary material, performed in a controlled 2-D setting, making it possible to easily visualize the behavior of  adversarial optimization when making use of different objective functions. 

\textbf{Equal Multiplier}\,\textemdash\,In studies that investigate adversarial examples, the perturbation generated with various sources is multiplied with a constant $\alpha$ and then added to the image in order to create an adversarial example. In our first experiment, we follow this common approach, using $\alpha = 5e-4$ for the equations given in Section~\ref{Experiments} to generate adversarial examples. However, in order to make a detailed analysis, instead of only analyzing the prediction, we also analyze the changes in magnitude of the gradient throughout the optimization, as well as the least amount of perturbation needed to change the prediction.

\textbf{Equal Perturbation}\,\textemdash\,
The gradient values produced depend on the loss function used, the subspace the optimized data point is located in, and the number of iterations executed thus far. As a result, the produced gradient values may be different in terms of their total magnitude. In order to make a fair comparison of optimization speed across different types of loss functions, we now diverge from the previous studies and adopt a dynamic multiplier technique with $\alpha = \beta \, /\sum_{i, j} |\mathbf{P}_{i, j}|$. This method adjusts the perturbation multiplier dynamically, so to make the added perturbation in each step equal (in terms of magnitude) across all methods for all iterations. The purpose of this experiment is to show the effectiveness of each perturbation produced using the aforementioned losses when the added perturbation is held constant. Under these circumstances, we use $\beta=5$ in order to calculate $\alpha$ dynamically.

\begin{table}[t]
\centering
\footnotesize
\caption{Mean and standard deviation, for the experiments \textit{``equal perturbation''} and \textit{``equal multiplier''},  of (1) the least number of iterations of added perturbation needed to change the prediction and (2) the time required to calculate the perturbation (measured on a single Titan-X GPU).}

\label{tbl:time_iter}
\centering
\begin{tabularx}{7cm}{ccccc}
\cline{2-5}
   \multicolumn{1}{c}{} 
 & \multicolumn{2}{c}{Equal Multiplier}  
 & \multicolumn{2}{c}{Equal Perturbation}\\
\cline{2-5}
   \multicolumn{1}{c}{} 
 &    \multicolumn{1}{c}{Iterations} 
 &    \multicolumn{1}{c}{Time (s)}
 &    \multicolumn{1}{c}{Iterations} 
 &    \multicolumn{1}{c}{Time (s)} \\
\cline{1-5}
CE & $114\pm71$ & $4.25\pm0.3$ &$55\pm28$ & $5.6\pm0.1$ \\
CE-Sign & $80\pm37$ & $4.05\pm0.2$ &$116\pm52$ & $5.6\pm0.1$ \\
Logit & $134\pm72$ & $4.19\pm0.2$ &$80\pm47$ & $5.7\pm0.1$ \\
M-Logit & $126\pm73$ & $4.30\pm0.27$ &$95\pm52$ & $5.8\pm0.1$ \\
\cline{1-5}
\end{tabularx}

\end{table}

\begin{figure}[t!]
\centering
\includegraphics[width=7cm,height=3.75cm]{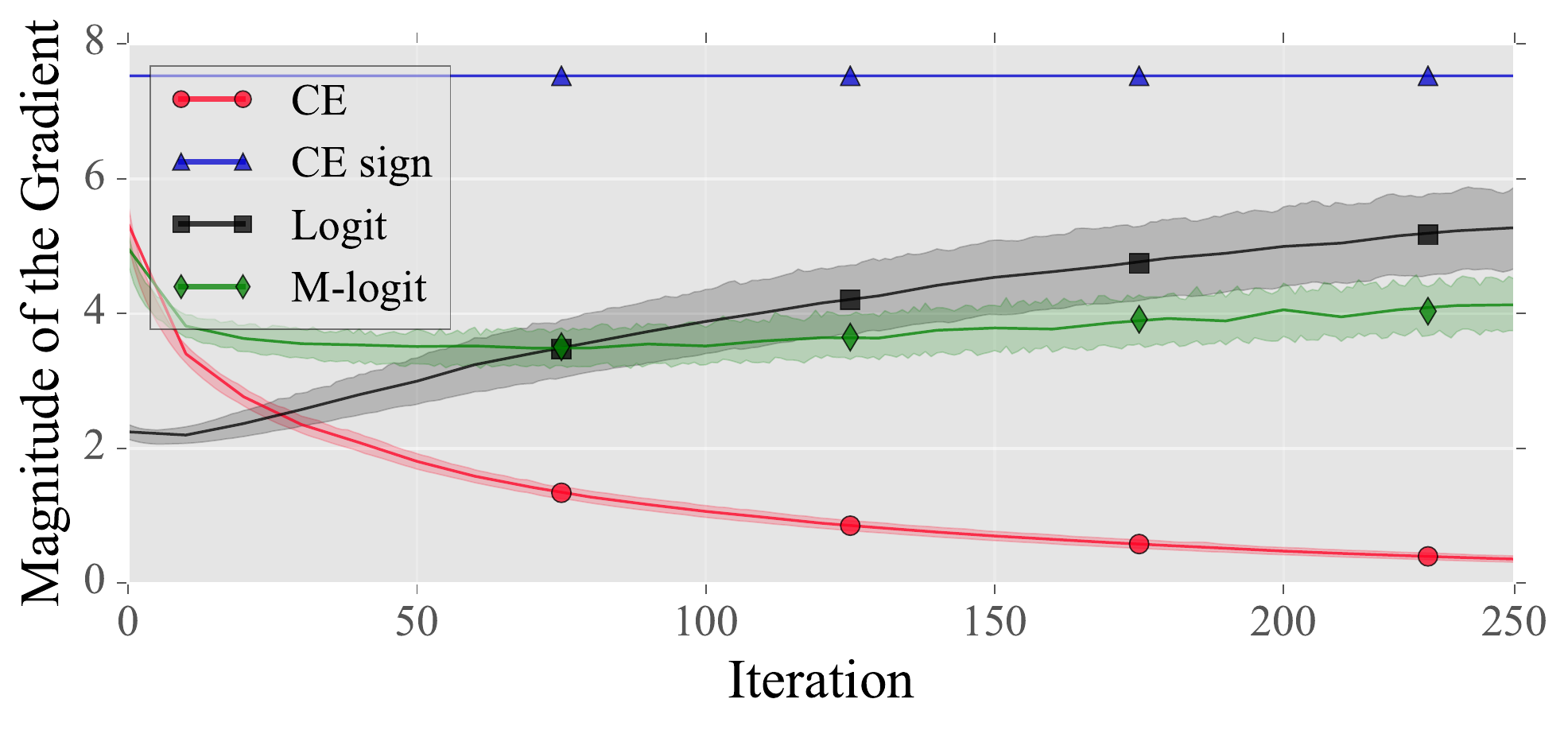}
\caption{Change in the magnitude of the gradient for the experiment \textit{``equal multiplier''}, for the data points presented in Fig.~\ref{fig:eqmul-opt}.}
\label{fig:fixmul-magnitude}

\end{figure}

\begin{figure*}[t]
\centering
\begin{tikzpicture}
\def\setxone{0}
\def\setyone{0}
\def\setytwo{\setyone + 1.6}
\def\setxtwo{\setxone + 1.6}
\node[inner sep=0pt] (c1) at (\setxone, \setyone)
    {\includegraphics[width=1.5cm,height=1.5cm]{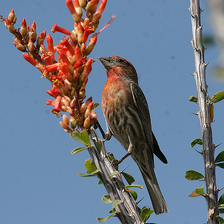}};
\node[inner sep=0pt] (t2) at (\setxone, \setytwo)
    {\includegraphics[width=1.5cm,height=1.5cm]{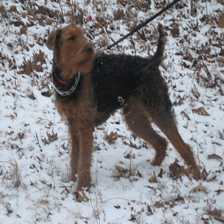}};
\node[inner sep=0pt] (c1) at (\setxtwo, \setyone)
    {\includegraphics[width=1.5cm,height=1.5cm]{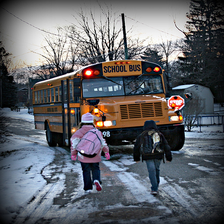}};
\node[inner sep=0pt] (t2) at (\setxtwo, \setytwo)
    {\includegraphics[width=1.5cm,height=1.5cm]{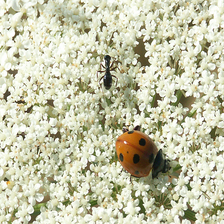}};
\node[align=center] at (0.8 , -1.05) {\footnotesize (a) Genuine image};

\def\setxthree{\setxtwo + 1.9}
\def\setxfour{\setxthree + 1.6}
\node[inner sep=0pt] (c1) at (\setxthree, \setyone)
    {\includegraphics[width=1.5cm,height=1.5cm]{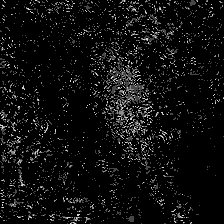}};
\node[inner sep=0pt] (t2) at (\setxthree, \setytwo)
    {\includegraphics[width=1.5cm,height=1.5cm]{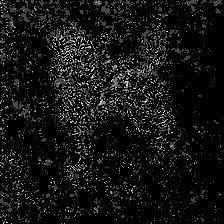}};
\node[inner sep=0pt] (c1) at (\setxfour, \setyone)
    {\includegraphics[width=1.5cm,height=1.5cm]{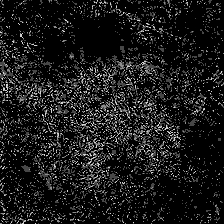}};
\node[inner sep=0pt] (t2) at (\setxfour, \setytwo)
    {\includegraphics[width=1.5cm,height=1.5cm]{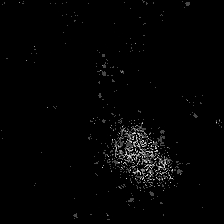}};
\node[align=center] at (\setxthree + 0.8 , -1.05) {\footnotesize (b) CE perturbation};

\def\setxfive{\setxfour + 1.9}
\def\setxsix{\setxfive + 1.6}
\node[inner sep=0pt] (c1) at (\setxfive, \setyone)
    {\includegraphics[width=1.5cm,height=1.5cm]{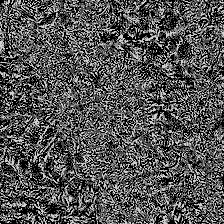}};
\node[inner sep=0pt] (t2) at (\setxfive, \setytwo)
    {\includegraphics[width=1.5cm,height=1.5cm]{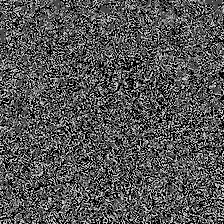}};
\node[inner sep=0pt] (c1) at (\setxsix, \setyone)
    {\includegraphics[width=1.5cm,height=1.5cm]{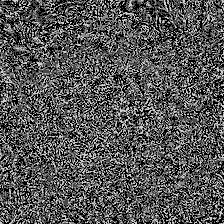}};
\node[inner sep=0pt] (t2) at (\setxsix, \setytwo)
    {\includegraphics[width=1.5cm,height=1.5cm]{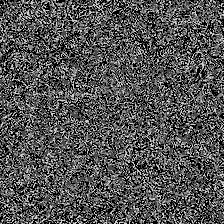}};
\node[align=center] at (\setxfive + 0.8 , -1.05) {\footnotesize (c) CE-Sign perturbation};

\def\setxseven{\setxsix + 1.9}
\def\setxeight{\setxseven + 1.6}
\node[inner sep=0pt] (c1) at (\setxseven, \setyone)
    {\includegraphics[width=1.5cm,height=1.5cm]{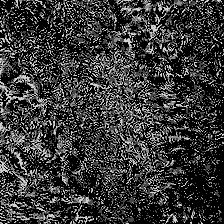}};
\node[inner sep=0pt] (t2) at (\setxseven, \setytwo)
    {\includegraphics[width=1.5cm,height=1.5cm]{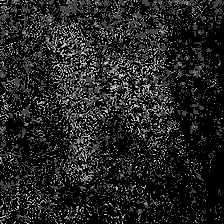}};
\node[inner sep=0pt] (c1) at (\setxeight, \setyone)
    {\includegraphics[width=1.5cm,height=1.5cm]{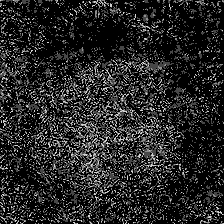}};
\node[inner sep=0pt] (t2) at (\setxeight, \setytwo)
    {\includegraphics[width=1.5cm,height=1.5cm]{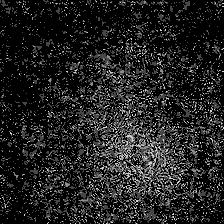}};
\node[align=center] at (\setxseven + 0.8 , -1.05) {\footnotesize (d) Logit perturbation};
    
\def\setxnine{\setxeight + 1.9}
\def\setxten{\setxnine + 1.6}
\node[inner sep=0pt] (c1) at (\setxnine, \setyone)
    {\includegraphics[width=1.5cm,height=1.5cm]{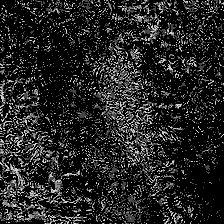}};
\node[inner sep=0pt] (t2) at (\setxnine, \setytwo)
    {\includegraphics[width=1.5cm,height=1.5cm]{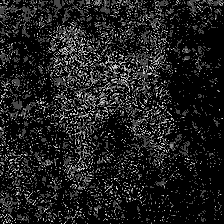}};
\node[inner sep=0pt] (c1) at (\setxten, \setyone)
    {\includegraphics[width=1.5cm,height=1.5cm]{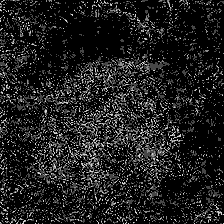}};
\node[inner sep=0pt] (t2) at (\setxten, \setytwo)
    {\includegraphics[width=1.5cm,height=1.5cm]{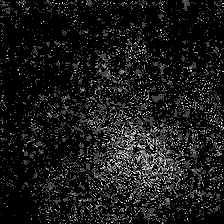}};
\node[align=center] at (\setxnine + 0.8 , -1.05) {\footnotesize (e) M-Logit perturbation};
\end{tikzpicture}
\caption{(a) Genuine images taken from the ImageNet dataset and the added perturbation illustrated in the form of saliency maps when the adversarial attacks are performed with (b) CE, (c) CE-Sign, (d) Logit, and (e) M-Logit losses.}
\label{fig:example_images}

\end{figure*}

\textbf{Experimental Results}\,\textemdash\,The experimental results obtained are displayed in Fig.~\ref{fig:eqmul-opt}, Fig.~\ref{fig:eqper-opt}, Fig.~\ref{fig:fixmul-magnitude}, and Table~\ref{tbl:time_iter}. Specifically, in Fig.~\ref{fig:eqmul-opt} and Fig.~\ref{fig:eqper-opt}, (a) shows the softmax output and (b) shows the logit prediction for both the initial class and the target class, for adversarial optimization using $250$ iterations, for both the experiment \textit{``equal multiplier''} and \textit{``equal perturbation''}. On the other hand, (c) in both Fig.~\ref{fig:eqmul-opt} and Fig.~\ref{fig:eqper-opt} visualizes the least amount of perturbation required in order to change the prediction to the target class, obtained once for all adversarial examples when their prediction is changed for the first time, with the amount of perturbation represented in the form of box plots for the respective experiments. The average number of iterations required to change the prediction of the data point under consideration and its standard deviation, as well as the total amount of time required to calculate the perturbations (for a total of $250$ iterations in terms of computer time), are listed in Table~\ref{tbl:time_iter}. Lastly, Fig.~\ref{fig:fixmul-magnitude} shows the change in gradient magnitude for the experiment \textit{``equal multiplier''}.

Based on these results, we make the following observations:

\textbullet\,\,When the perturbation multiplier is held the same, the magnitude of the produced gradient becomes substantially different for each method, making this kind of experiments unsuitable for analyzing how \textit{fast} or \textit{slow} a method is.

\vspace{0.25em}

\textbullet\,\,The misconception that the CE-sign loss, as used by IFGS and FGS, generates adversarial examples \textit{faster} than the other methods~\cite{CW_Attack} stems from the fact that taking the signature naturally boosts the gradient obtained from the model. This is shown in Fig.~\ref{fig:fixmul-magnitude}, where the perturbation added by the CE-sign loss is significantly larger than the perturbation added by other approaches. However, when the perturbation is held constant, the CE-sign becomes the least effective loss and creates adversarial examples the slowest in terms of the number of iterations needed.
\vspace{0.25em}

\textbullet\,\,The CE loss, whose usage is criticized in the works of \cite{CW_Attack,DBLP:journals/corr/CarliniW17}, on average, not only creates adversarial examples the fastest when the perturbation is held constant, but also does so with the least amount of total $L_2$ perturbation compared to other losses.

\vspace{0.25em}

\textbullet\,\,Even when the perturbation multiplier is held the same across all iterations, the gradient generated by the CE loss decreases as the optimization continues (see Fig.~\ref{fig:fixmul-magnitude}), confirming the practicality of Theorem~\ref{thm:1}.

\vspace{0.25em}

\textbullet\,\,Both the CE and the CE-sign loss have a limited optimization space, as observed in Fig.~\ref{fig:eqmul-line-2} and Fig.~\ref{fig:eqper-line-2}. This explains why the losses that rely on CE are not able to create \textit{strong} adversarial examples.

\vspace{0.25em}

\textbullet\,\,Although, on average, the M-logit loss creates adversarial examples slower (95 iterations) than the logit loss (80 iterations), we observe that its total added perturbation is less than that of the logit loss. This means that more iterations of added perturbation does not necessarily equate to a larger total perturbation, further showing the effectiveness of the M-logit loss, as used by \cite{CW_Attack}.

\vspace{0.25em}

\textbullet\,\,Even if the M-logit loss indeed seems to be the slowest method, which confirms the criticism of \cite{Goodfellow:2018:MML:3234519.3134599}, the computational time difference between each method and M-logit loss is not substantial: the M-logit loss is, on average, only $3\%$ slower than the fastest method. What makes CW extremely slow is its extensive search for the most effective perturbation, as described in \cite{CW_Attack}, and not its underlying loss function.

\vspace{0.25em}

\textbullet\,\,Even though we observe a correlation between the total amount of added perturbation in terms of $L_2$ and $L_{\infty}$ distances, the results for the CE-sign loss are vastly different from others. This is because taking the signature gets rid of changes with large magnitude, which will especially impact substantial changes. This results in significantly smaller perturbations in terms of $L_{\infty}$ distance. 

{\color{royalazure}
\subsection{Identification of Adversarial Perturbations}

One of the first methods tested to prevent adversarial attacks was the method of adversarial retraining~\cite{grosse2017statistical}. This method is performed by retraining the model under consideration with adversarial examples while adding an additional ($(M+1)th$) adversarial class to the set of original classes. Adversarial retraining was later extended by~\cite{gong2017adversarial}, with the authors training a model in a binary fashion in order to obtain a discriminator that distinguishes genuine images from adversarial ones. Because this experiment was performed with a limited number of images and on datasets that are not fit to study adversarial attacks (due to limitations in terms of (1) color channels (MNIST~\cite{lecun1998gradient}), (2) image sizes (CIFAR~\cite{CIFAR}), or (3) the total number of images available), the method of adversarial retraining was later labeled as an ineffective defense mechanism~\cite{DBLP:journals/corr/CarliniW17}.

In this experiment, we revisit the idea of adversarial retraining. For each attack described in Section~\ref{Experimental Setup}, we generate $50,000$ adversarial examples using a pretrained ResNet-50 network, taking initial images from the validation set of the ImageNet dataset. This artificial dataset of $200,000$ adversarial images is then utilized to train AlexNet~\cite{Alexnet}, VGG-16~\cite{VGG}, and ResNet-18/34/152~\cite{resnet} in order to analyze whether these neural networks can distinguish genuine images from adversarial ones. In the first part of this experiment, we analyze the detectability of adversarial examples for each attack separately (i.e., models trained with genuine images and adversarial examples generated with a single type of attack at a time). In the second part, we incorporate adversarial examples generated with all four types of attacks, analyzing whether this approach is effective as a defense mechanism and which types of attacks are harder to detect.

A detailed description of the parameters used during the training of these models, as well as dataset splits, can be found in the supplementary material.

\begin{table}[t]
\caption{Accuracy (genuine / adversarial (overall)) of binary classification between genuine images and adversarial examples, and where the adversarial examples have been generated by the attacks listed in the first column for the given models. Each entry containing three values represents the result of an adversarial retraining. Architectures are ordered from the left to the right in terms of ascending convolutional layer complexity.}

\label{tbl:adv_retraining1}
\scalebox{0.90}{
\footnotesize
\begin{tabularx}{4cm}{cccccc}
\cmidrule[1pt]{1-6}
   \multicolumn{1}{c}{Attack} 
 &    \multicolumn{1}{c}{AlexNet} 
 &    \multicolumn{1}{c}{ResNet-18}
 &    \multicolumn{1}{c}{VGG-16}
 &    \multicolumn{1}{c}{ResNet-34} 
 &    \multicolumn{1}{c}{ResNet-152} \\
\cmidrule[0.5pt]{1-6}
\multirow{ 2}{*}{\shortstack{CE}} & $88\%\,30\%$ & $80\%\,79\%$ & $94\%\,90\%$ & $92\%\,90\%$ & $96\%\,94\%$ \\
~ & $(59\%)$ & $(79\%)$ & $(92\%)$ & $(91\%)$ & $(95\%)$\\
\cmidrule[0.5pt]{2-6}
\multirow{ 2}{*}{\shortstack{CE-Sign}} & $90\%\,94\%$ & $97\%\,96\%$ & $98\%\,99\%$ & $99\%\,99\%$ & $99\%\,99\%$ \\
~ & $(92\%)$ & $(96\%)$ & $(98\%)$ & $(99\%)$ & $(99\%)$\\
\cmidrule[0.5pt]{2-6}
\multirow{ 2}{*}{\shortstack{Logit}} & $94\%\,94\%$ & $95\%\,96\%$ & $98\%\,97\%$ & $98\%\,99\%$ & $99\%\,99\%$ \\
~ & $(94\%)$ & $(95\%)$ & $(97\%)$ & $(98\%)$ & $(99\%)$\\
\cmidrule[0.5pt]{2-6}
\multirow{ 2}{*}{\shortstack{M-Logit}} & $74\%\,75\%$ & $90\%\,91\%$ & $95\%\,96\%$ & $96\%\,96\%$ & $98\%\,97\%$ \\
~ & $(74\%)$ & $(90\%)$ & $(95\%)$ & $(96\%)$ & $(97\%)$\\
 \cmidrule[1pt]{1-6}
\end{tabularx}
}

\end{table}

\begin{table}[t]
\caption{
Breakdown of classification accuracy for genuine images and adversarial examples when four types of attacks are incorporated. Each column represents the result of an adversarial retraining. Architectures are ordered from the left to the right in terms of ascending convolutional layer complexity.}

\label{tbl:adv_retraining2}
\scalebox{0.90}{
\footnotesize
\begin{tabularx}{4cm}{cccccc}
\cmidrule[1pt]{1-6}
   \multicolumn{1}{c}{Attack}
 &    \multicolumn{1}{c}{AlexNet} 
 &    \multicolumn{1}{c}{ResNet-18}
 &    \multicolumn{1}{c}{VGG-16}
 &    \multicolumn{1}{c}{ResNet-34} 
 &    \multicolumn{1}{c}{ResNet-152} \\
 \cmidrule[0.5pt]{1-6}
CE & $53\%$ & $83\%$ & $95\%$ & $92\%$ & $93\%$  \\
\cmidrule[0.25pt]{2-6}
CE-Sign & $76\%$ & $97\%$ & $99\%$ & $98\%$ & $99\%$  \\
\cmidrule[0.25pt]{2-6}
Logit & $71\%$ & $95\%$ & $98\%$ & $97\%$ & $99\%$  \\
\cmidrule[0.25pt]{2-6}
M-Logit & $52\%$ & $84\%$ & $96\%$ & $93\%$ & $94\%$  \\
\cmidrule[0.5pt]{1-6}
\multirow{2}{*}{\shortstack{Adversarial\\Examples}}  & \multirow{2}{*}{$64\%$} & \multirow{2}{*}{$90\%$}  & \multirow{2}{*}{$97\%$}  & \multirow{2}{*}{$95\%$} & \multirow{2}{*}{$97\%$}  \\
\\
\cmidrule[0.25pt]{2-6}
\multirow{2}{*}{\shortstack{Genuine\\Images}}  & \multirow{2}{*}{$81\%$} & \multirow{2}{*}{$94\%$}  & \multirow{2}{*}{$93\%$}  & \multirow{2}{*}{$95\%$} & \multirow{2}{*}{$99\%$}  \\
\\
\cmidrule[0.25pt]{1-6}
\multirow{2}{*}{\shortstack{Overall\\Accuracy}}  & \multirow{2}{*}{$72\%$} & \multirow{2}{*}{$92\%$}  & \multirow{2}{*}{$95\%$}  & \multirow{2}{*}{$95\%$} & \multirow{2}{*}{$98\%$}  \\
\\
\cmidrule[1pt]{1-6}
\end{tabularx}
}

\end{table}

\textbf{Experimental Results}\,\textemdash\,Experimental results obtained for adversarial retraining can be found in Table~\ref{tbl:adv_retraining1} and Table~\ref{tbl:adv_retraining2}, with the former representing the classification accuracy when models are retrained with genuine images and only one type of adversarial examples, and with the latter showing the classification accuracy when the adversarial retraining is done with genuine images combined with all types of adversarial examples. Furthermore, in Table~\ref{tbl:adv_retraining2}, next to providing the classification accuracy of adversarial examples for each type of attack, we also provide aggregate results in order to present the detectability per attack type. Given the experimental results shown in Table~\ref{tbl:adv_retraining1} and Table~\ref{tbl:adv_retraining2}, our findings can be summarized as follows:

\textbullet\,\,Adversarial retraining requires massive amounts of data. The reason why we had to generate such a large artificial dataset was that any adversarial retraining effort conducted with less than this amount of data produced inconsistent results. This observation is also made by \cite{gong2017adversarial}, where the classifier trained achieved an accuracy of either $0\%$ or $100\%$.

\vspace{0.25em}

\textbullet\,\,Different from the study of~\cite{DBLP:journals/corr/CarliniW17}, we find adversarial retraining to be an effective method for detecting adversarial examples that have been generated using ImageNet, for which the perturbation patterns obtained are more distinctive than the perturbation patterns obtained for datasets with smaller image sizes (see Fig.~\ref{fig:example_images} for perturbation examples).

\vspace{0.25em}

\textbullet\,\,We observe a clear correlation between the capability of the model (in terms of convolutional layer complexity) and the ability of discriminating genuine images from adversarial ones, an observation also made by~\cite{PGD_attack}.

\vspace{0.25em}

\textbullet\,\,Confirming~\cite{gong2017adversarial}, we also find that adversarial examples, as generated with CE-Sign, are the easiest to detect in comparison to other types of adversarial examples. In addition, we find that adversarial examples generated with CE and M-Logit are harder to detect when training incorporates all adversarial examples.

\vspace{0.25em}

\textbullet\,\,It might seem like CE is a decent baseline approach to generate adversarial examples. However, due to its constrained optimization space and limited perturbation generation capacity (as previously discussed in Section~\ref{Perturbation Analysis}), adversarial examples generated with CE are less likely to transfer to other models, and are more susceptible to defense mechanisms that use input transformations such as total variation or blurring \cite{DBLP:journals/corr/CarliniW17}.

}

\section{Conclusions and Future Research} \label{Conclusion}

In this paper, we first generalized a number of popular methods for adversarial example generation, making their objective functions mathematically comparable in terms of the way perturbation is generated. That way, we were able to establish that each of these objective functions falls into one out of two categories: the ones that use cross-entropy loss and the ones that use logit loss. Next, we demonstrated that the gradient of the cross-entropy loss can be approximated differently for three prediction subspaces. We established both mathematically and empirically that using the CE loss --- or any other loss function that, at its core, relies on CE --- has only a limited target subspace when creating adversarial examples, as compared to the usage of the logit loss.

Our results show that, when the cross-entropy loss is used as a baseline loss function to generate adversarial examples, it is not possible to create adversarial examples whose logit output exceeds a certain limit, due to mathematical constraints, whereas it is possible to do so when using the logit loss. However, this does not mean that the use of CE is always detrimental to the process of adversarial example generation. On the contrary, compared to all other methods studied in this paper, the use of CE creates adversarial examples the fastest and with the least amount of $L_2$ perturbation. 
{\color{royalazure}
Next, we observed that neural networks are indeed able to identify adversarial perturbation patterns as generated by individual attacks and that the method of adversarial retraining can be used to differentiate genuine images from their adversarial counterparts on ImageNet, given that the perturbation patterns obtained for this dataset are more meaningful than the perturbation patterns obtained for datasets with smaller image sizes.}

With this study, we aim at guiding future research efforts that focus on designing novel methods for adversarial example generation, showing the impact of the selected baseline loss function on the behavior of adversarial optimization. We also aim at guiding future research efforts that focus on constructing novel defenses against adversarial examples by exposing the optimization space limitation of the cross-entropy loss, which in turn enables defenses that focus on the different subspaces we identified. {\color{royalazure} Lastly, in the case of adversarial retraining, and from a computer vision perspective, it may be of interest to analyze what exactly the models are learning in order to achieve a better understanding of the adversarial perturbation patterns.}

\bibliographystyle{IEEEtran}
\bibliography{2019_04}

\onecolumn
\clearpage

\section{Omitted Proofs}

\textit{
\textbf{Theorem 5.1}
Let $\mathbf{y} =g(\theta, \mathbf{X})$ be a neural network, represented by a differentiable function $g$ that maps an input $\mathbf{X}$ to an output vector $\mathbf{y}$. Furthermore, assume that adversarial examples are generated using the cross-entropy function $J(g(\theta, \mathbf{X})_{c})$. \newline\newline
Based on the subpaces described in Definition (5.1) in the main text, the gradient $\nabla_x J(g(\theta, \mathbf{X})_{c})$ of the cross-entropy function can be approximated as follows:
\begin{itemize}
\item $\forall \mathbf{X}_1 \in D_1,$
\begin{align} 
\lim_{\mathbf{X} \to \mathbf{X}_{1}} \nabla_{x} J(g(\theta, \mathbf{X})_{c}) = \nabla_{x} g(\theta , \mathbf{X}_1)_{r} - \nabla_{x} g(\theta , \mathbf{X}_1)_{c} \,.
\end{align}
\item $\forall \mathbf{X}_2 \in D_2,$
\begin{align} 
\lim_{\mathbf{X} \to \mathbf{X}_{2}} & \nabla_{x} J(g(\theta, \mathbf{X})_{c}) = 
\sum_{m \in I\setminus \{c\}} \beta_m  \Big(\nabla_{x} g(\theta , \mathbf{X}_2)_{m} 
 - \nabla_{x} g(\theta , \mathbf{X}_2)_{c}\Big)\,, 
\end{align}
with constants $\beta_1,\ldots,\beta_M$ subject to $\sum_{m \in I\setminus \{c\}} | \beta_m | < 1$, where $I = \{1, 2, \ldots, M\}$.
\newcommand\inlineeqno{\stepcounter{equation}\ (\theequation)} 
\item $\forall \mathbf{X}_3 \in D_3,$ \,\quad $\lim_{\mathbf{X} \to \mathbf{X}_{3}} \nabla_{x} J(g(\theta, \mathbf{X})_{c}) =  0\,. \,\qquad\qquad\, \inlineeqno$
\end{itemize}
}
\begin{proof}[Proof of Theorem 5.1]
Recall the formulae of the cross-entropy function $J$:
\begin{align}
J(g(\theta, \mathbf{X})_{c}) = - \log \left( \dfrac{ e^{g( \theta, \mathbf{X})_{c}} }{ \sum_{m=1}^{M} e^{g( \theta , \mathbf{X})_{m}}  } \right) \,,
\end{align}
where $c \in [1, M]$ is the target label among $M$ classes. We can rewrite the cross-entropy function as
\begin{align}
J(g(\mathbf{X})_{c}, \theta) = - g( \theta, \mathbf{X})_{c} + \log\left( \sum_{m=1}^{M} e^{g( \theta , \mathbf{X})_{m}} \right) \,.
\end{align}
To gain insight into the behavior of the cross-entropy loss, we will, given Equation 5 in the main text, analyze its gradient 
\begin{align}
\nabla_{x} J(g(\theta, \mathbf{X})_{c}) &= \nabla_{x} \left( - g( \theta, \mathbf{X})_{c} + \log \left( \sum_{m=1}^{M} e^{g( \theta , \mathbf{X})_{m}} \right)\right) \,, \\
 &= - \nabla_{x} g(\theta , \mathbf{X})_{c} + 
\frac{\displaystyle \sum_{m=1}^{M} e^{g(\theta , \mathbf{X})_{m}} \, \nabla_{x} g(\theta , \mathbf{X})_{m}  }{ \displaystyle \sum_{m=1}^{M} e^{g(\theta , \mathbf{X})_{m}} } \,.
\label{eq:ce_fin}
\end{align}
Let us now consider Equation~(\ref{eq:ce_fin}) for the different subspaces. Denote by $I = \{1, 2, \ldots, M\}$ the set of class indices.
\begin{itemize}
    \item When in subspace $D_1$, each data point is classified with high confidence as belonging to some class $r \in I \setminus\{c\}$. Then, $\forall \mathbf{X}_1 \in D_1$, we have $e^{g(\theta , \mathbf{X}_1)_{r}} \gg e^{g(\theta , \mathbf{X}_1)_{m \in I \setminus \{r\}}}$. Hence,
    \begin{align}
    \lim_{\mathbf{X} \to \mathbf{X}_{1}} \frac{ \displaystyle \sum_{m=1}^{M} e^{g(\theta , \mathbf{X})_{m}} \, \nabla_{x} g(\theta , \mathbf{X})_{m}  }{\displaystyle \sum_{m=1}^{M} e^{g(\theta , \mathbf{X})_{m}} }  =   \nabla_{x} g(\theta , \mathbf{X}_1)_{r} \,,
    \end{align}
    which means that
    \begin{align}
    \lim_{\mathbf{X} \to \mathbf{X}_{1}} \nabla_{x} J(g(\theta, \mathbf{X})_{c}) =   \nabla_{x} g(\theta , \mathbf{X}_1)_{r}  -  \nabla_{x} g(\theta , \mathbf{X}_1)_{c} \,.
    \end{align}
    \item While in subspace $D_3$, each data point is classified, with high confidence, as belonging to the target class: $\forall \mathbf{X}_3 \in D_3$, $e^{g(\theta , \mathbf{X}_1)_{c}} \gg e^{g(\theta , \mathbf{X}_1)_{m \in I \setminus \{c\}}}$. Thus,
    \begin{align}
    \lim_{\mathbf{X} \to \mathbf{X}_{3}} \frac{ \displaystyle \sum_{m=1}^{M} e^{g(\theta , \mathbf{X})_{m}} \, \nabla_{x} g(\theta , \mathbf{X})_{m}  }{ \displaystyle \sum_{m=1}^{M} e^{g(\theta , \mathbf{X})_{m}} }  =   \nabla_{x} g(\theta , \mathbf{X}_3)_{c} \,.
    \end{align}
    Therefore
    \begin{align}
    & \lim_{\mathbf{X} \to \mathbf{X}_{3}} \nabla_{x} J(g(\theta, \mathbf{X})_{c}) =  0\,.
    \label{eq:zero}
    \end{align}
    \item Finally, for subspace $D_2$, we will first continue rewriting Equation~(\ref{eq:ce_fin}):
    \begin{align}
    & \nabla_{x} J(g(\theta, \mathbf{X})_{c}) =\\
    &- \nabla_{x} g(\theta , \mathbf{X})_{c} + 
    \frac{ \displaystyle \sum_{m\in I} e^{g(\theta , \mathbf{X})_{m}} \, \nabla_{x} g(\theta , \mathbf{X})_{m}  }{ \displaystyle \sum_{m \in I} e^{g(\theta , \mathbf{X})_{m}} } \,,  \\
    & = - \frac{ \displaystyle \sum_{m\in I} e^{g(\theta , \mathbf{X})_{m}} \, \nabla_{x} g(\theta , \mathbf{X})_{c}   }{ \displaystyle \sum_{m \in I} e^{g(\theta , \mathbf{X})_{m}} } 
    + \frac{ \displaystyle \sum_{m\in I} e^{g(\theta , \mathbf{X})_{m}} \, \nabla_{x} g(\theta , \mathbf{X})_{m}  }{ \displaystyle \sum_{m \in I} e^{g(\theta , \mathbf{X})_{m}} } \,,  \\
    & = \frac{ \displaystyle \sum_{m\in I} e^{g(\theta , \mathbf{X})_{m}} \, \Big(\nabla_{x} g(\theta , \mathbf{X})_{m} - \nabla_{x} g(\theta , \mathbf{X})_{c} \Big)  }{ \displaystyle \sum_{m \in I} e^{g(\theta , \mathbf{X})_{m}} } \,.
    \label{eq:D2_1}
    \end{align}
\end{itemize}
For $m=c$, $e^{g(\theta , \mathbf{X})_{m}} \, (\nabla_{x} g(\theta , \mathbf{X})_{m} - \nabla_{x} g(\theta , \mathbf{X})_{c}) =0 $. This means that the case $m=c$ does not contribute to the sum and that we can write Equation~(\ref{eq:D2_1}) as:
\begin{align}
\nabla_{x} J(g(\theta, \mathbf{X})_{c}) &=\frac{\displaystyle \sum_{m\in I\setminus\{c\}} e^{g(\theta , \mathbf{X})_{m}} \, \Big(\nabla_{x} g(\theta , \mathbf{X})_{m} - \nabla_{x} g(\theta , \mathbf{X})_{c}\Big)  }{ \displaystyle \sum_{m \in I} e^{g(\theta , \mathbf{X})_{m}} } \,.
\end{align}

Trivially, $\forall m \in I$, $\beta_m := \frac{e^{g(\theta , \mathbf{X})_{m}}   }{  \sum_{m \in I} e^{g(\theta , \mathbf{X})_{m}} } < 1$ since we know that $\forall \mathbf{X}_2 \in D_2$. We can then represent the gradient as
\begin{align}
\lim_{\mathbf{X} \to \mathbf{X}_{2}} \nabla_{x} J(g(\theta, \mathbf{X})_{c}) = \displaystyle \sum_{m \in I\setminus \{c\}} \beta_m \Big(\nabla_{x} g(\theta , \mathbf{X}_2)_{m} - \nabla_{x} g(\theta , \mathbf{X}_2)_{c}\Big)
\end{align}
with $\displaystyle \sum_{m \in I\setminus \{c\}} | \beta_m | < 1$.
\end{proof}

\clearpage
\begin{figure*}[ht]
\centering
\begin{subfigure}{0.21\textwidth}
\includegraphics[width=4.1cm]{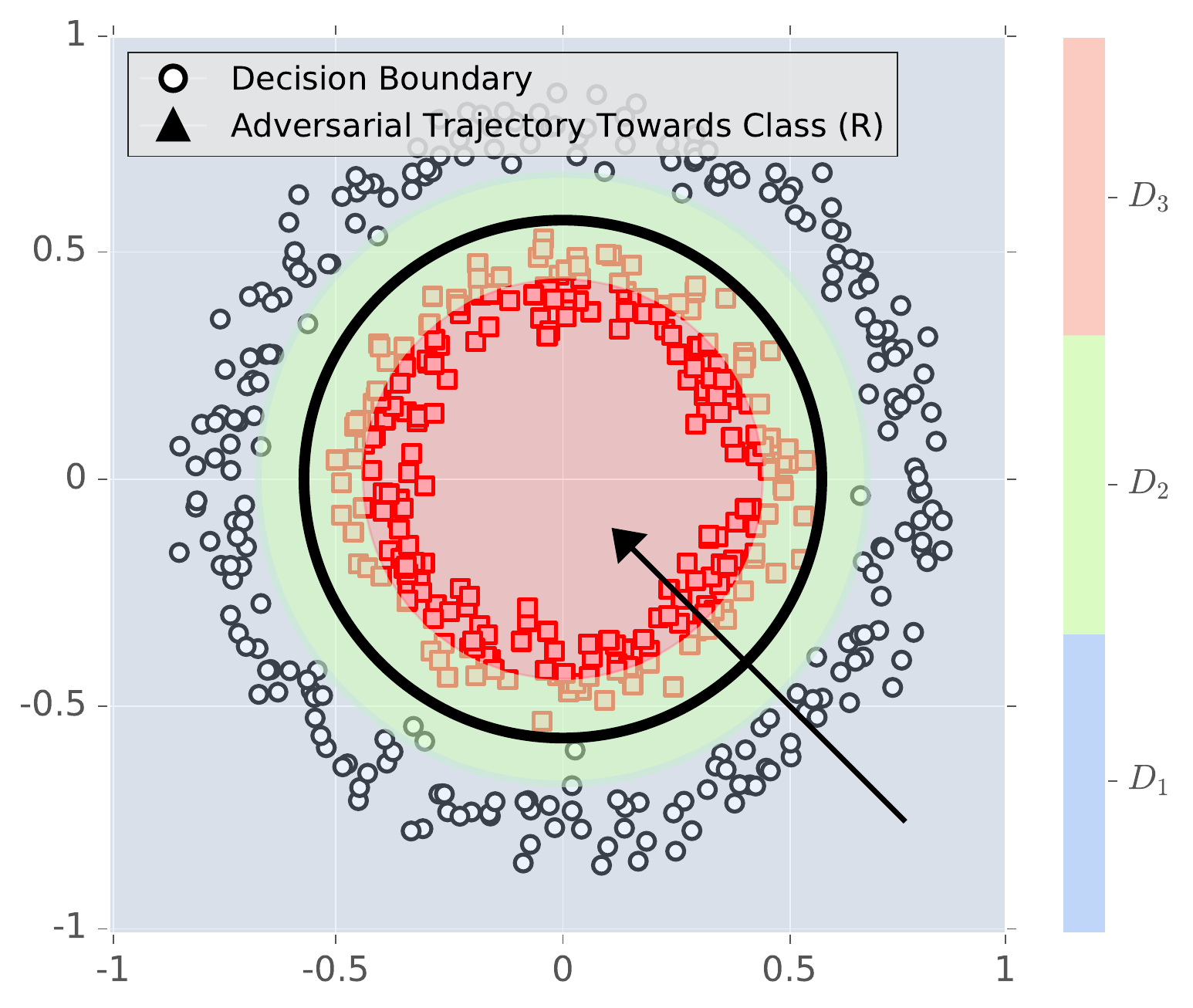}
\caption{Experimental setting}
\label{opt-1}
\end{subfigure}
\begin{subfigure}{0.21\textwidth}
\includegraphics[width=4.1cm]{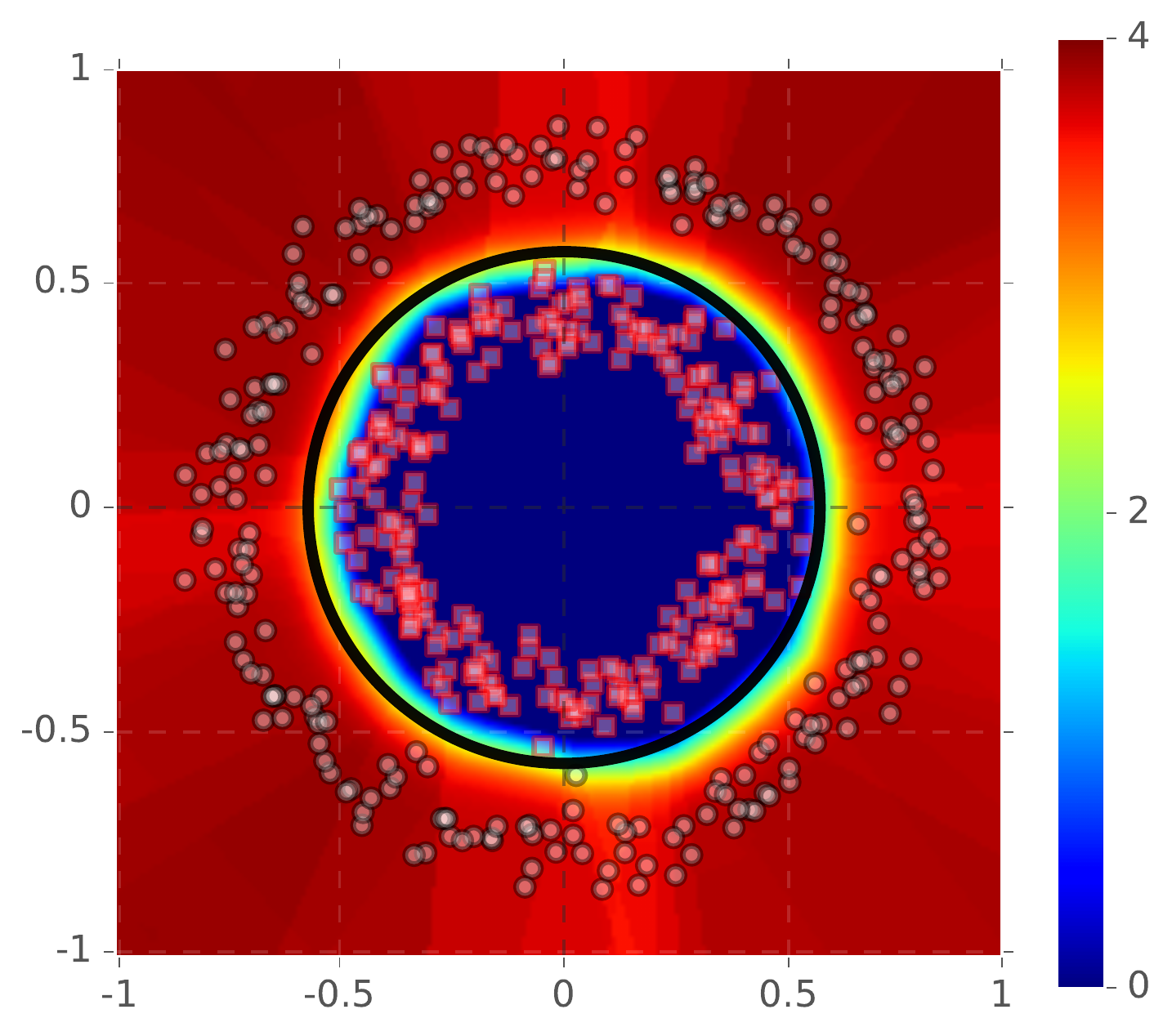}
\caption{CE loss}
\label{opt-2}
\end{subfigure}
\begin{subfigure}{0.21\textwidth}
\includegraphics[width=4.1cm]{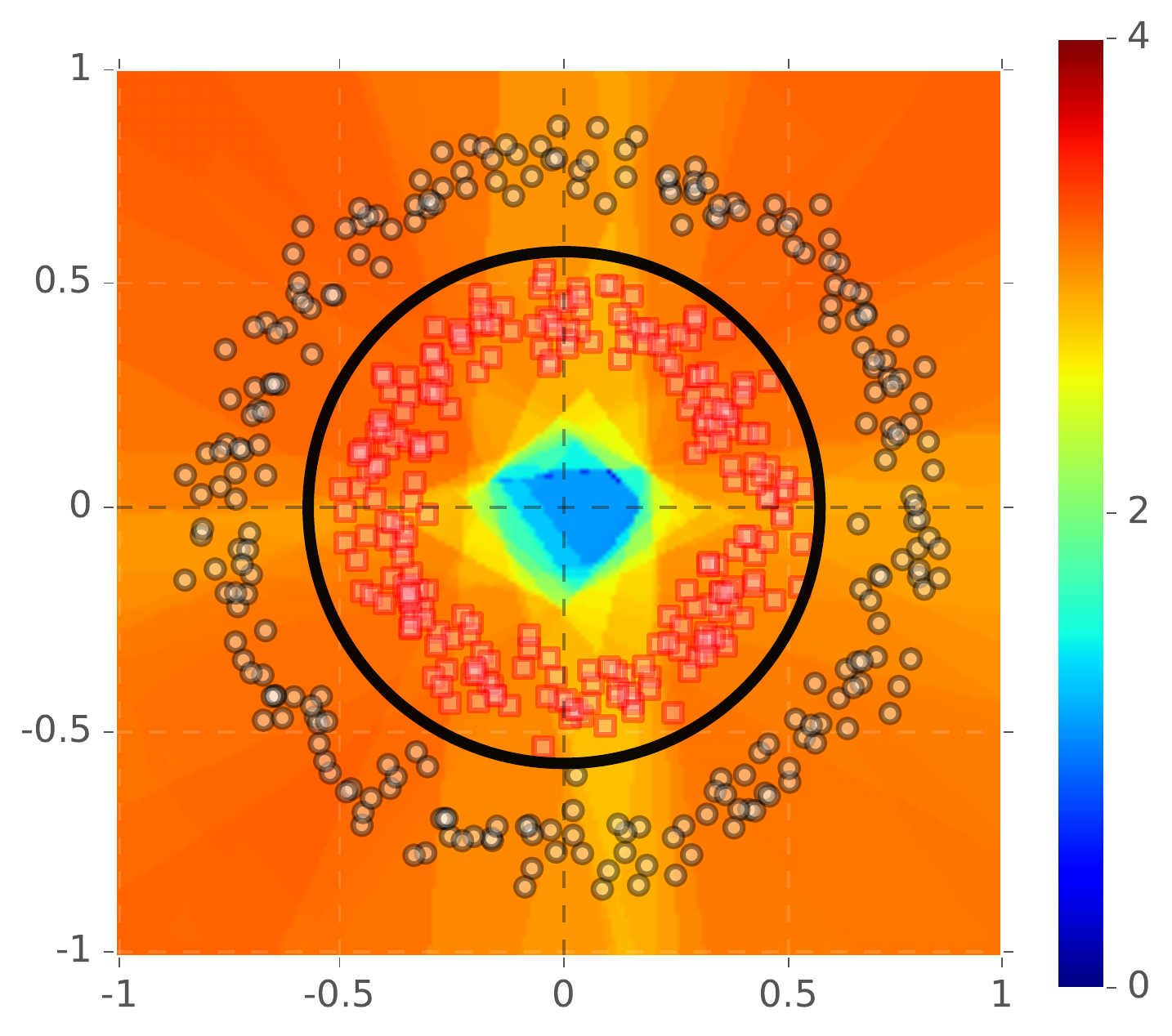}
\caption{Logit loss}
\label{opt-3}
\end{subfigure}
\begin{subfigure}{0.21\textwidth}
\includegraphics[width=4.1cm]{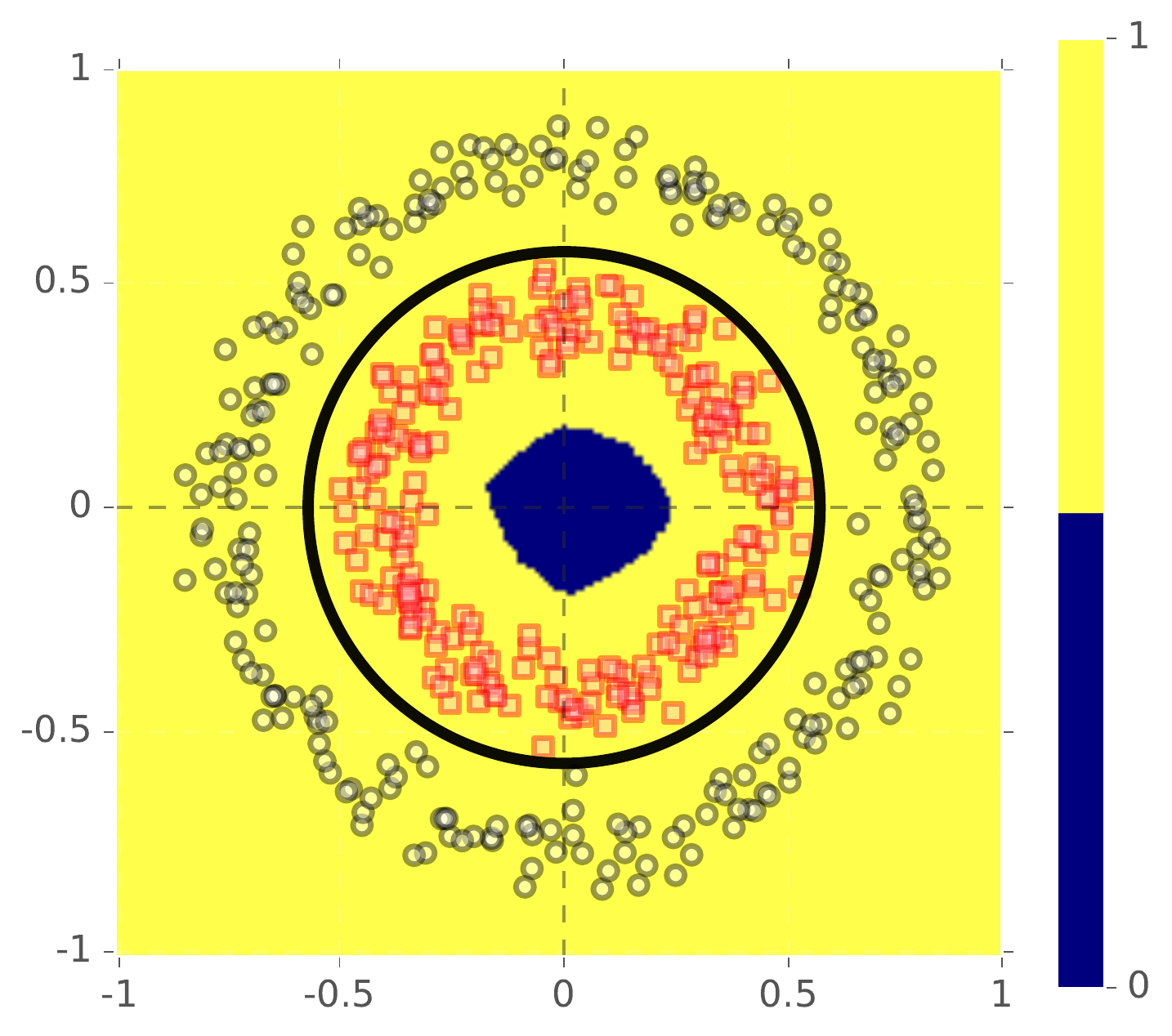}
\caption{CE sign loss}
\label{opt-4}
\end{subfigure}
\label{fig:opt-simulation}
\caption{(a) A classification problem consisting of two circular distributions with the same center $(0,0)$ but different radii. The data are bounded by $(x,y) \in [-1, 1]^2$. An adversarial trajectory is defined as moving a data point from the outer class to the inner class across the decision boundary. Heat maps characterizing the adversarial optimization are given for the classification problem presented in (a), generated using (b) the CE loss, $ \log \left(\sum | \nabla_{x} J(g(\theta, \mathbf{X})_{(R)})|\right)$; (c) the logit loss, $ \log \left(\sum | \nabla_{x} g(\theta, \mathbf{X})_{(R)}|\right)$; and (d) the CE sign loss, $\text{sign}\left( \sum | \nabla_{x} J(g(\theta, \mathbf{X})_{(R)}))|\right)$.}
\end{figure*}

\section{Additional Experiments and Experimental Details}
\textbf{Experiments on 2-D}\,\textemdash\,To visually demonstrate how adversarial optimization behaves for different objective functions, we use the setting of adversarial example generation illustrated in Figure~\ref{opt-1}. This experiment deals with a 2-D two-class classification problem where the data are sampled using the function \textit{make\_circles} of the SciPy library \cite{scipy_cite}, obtaining two circular data motifs with the same mean $(0,0)$, but different radii. To this problem, we apply a neural network with a single hidden layer that contains $50$ neurons followed by a rectifier activation~\cite{glorot2011deep}. Our model takes the 2-D coordinates $(x, y)$ as input and maps each point to one out of two target classes (R)ed and (B)lack. The circular decision boundary of this model is drawn in black in Figure~\ref{opt-1}. Under these settings, the adversarial example generation process is defined as moving a point from the outer class (B) to the inner class (R), the latter thus acting as the \textit{target class}.

The potency of the loss is visualized in the form of heat maps. Since the direction of optimization is different for different points due to the circularity of the distributions, we use the absolute magnitude of the gradient in these heat maps. Figure~\ref{opt-2}, Figure~\ref{opt-3}, and Figure~\ref{opt-4} represents the magnitude of the gradients generated with CE loss, CE sign loss, and logit loss for the points $(x,y) \in [-1, 1]^2$, respectively. For each figure, we provide summarizing observations below:
\begin{itemize}
    \item Figure~\ref{opt-2}\,\textemdash\,The limited optimization space of the CE loss can be clearly observed, with the magnitude of the gradient becoming zero as soon as data points in the target subspace $D_3$ are selected.
    \item Figure~\ref{opt-3}\,\textemdash\,When the logit loss is used, the gradient exists for all points $(x,y) \in [-1, 1]^2$. Yet, for the logit loss, the magnitude of the gradient for data points in $D_1$ is less than the magnitude for the same data points when CE loss is selected, which shows the benefit in terms of optimization speed when using CE loss over logit loss for the data points that lie in $D_1$. 
    \item Figure~\ref{opt-4}\,\textemdash\,The area that contains non-zero gradients in the target subspace $D_3$ has increased compared to Figure~\ref{opt-2}. However, there still exists an area at the center of the graph where the magnitude of the gradient is exactly zero.
\end{itemize}

\textbf{Adversarial Retraining Details}\,\textemdash\,In what follows, we provide details about the data set and the training methodology used for the experiments described in Section~5.2 of the main text.
\begin{itemize}
    \item Architecture\,\textemdash\,For the adversarial retraining experiments, we adopt AlexNet~\cite{Alexnet}, VGG-16~\cite{VGG}, and ResNet-18/34/152~\cite{resnet}, performing weight initialization using the models available in the PyTorch library~\cite{paszke2017automatic}, pretrained on the ImageNet training set. We then replace the final linear layer with another newly initialized layer that has two class outputs (one for genuine images and one for adversarial images). The weights for this layer are initialized with the initialization method provided in ~\cite{resnet}.
    \item Data set\,\textemdash\,We generate $50,000$ adversarial examples for each attack using a pretrained ResNet-50 network in white-box settings, with initial images for the adversarial attacks taken from the images in the ImageNet validation set.
    \begin{itemize}
        \item Training on a single type of adversarial examples\,\textemdash\,The results provided in Table~2, as available in the main text, were obtained by training the aforementioned architectures in a binary fashion, where the set of adversarial examples only contains those adversarial examples generated by a single attack (provided in the first column of Table~2). For training purposes, we use $47,500$ genuine images available in the ImageNet validation set and their $47,500$ adversarial counterparts. The remaining $5,000$ images (half genuine, half adversarial) are used as a test set.
        \item Training on multiple types of adversarial examples\,\textemdash\,The results provided in Table~3, as available in the main text, were obtained by training the aforementioned architectures in a binary fashion, where all adversarial examples generated by all attacks are labeled as the adversarial class. We use $50,000$ genuine images available in the ImageNet validation set and $200,000$ adversarial examples, with the latter generated from the aforementioned genuine images by four types of attacks. Due to the imbalance between both classes, we incorporate $150,000$ extra genuine images taken from the extended data set ImageNet-10K \cite{ImageNet10k}. The reason for not using any images from the training set of ImageNet is that the models we apply are already pretrained on the training set of ImageNet. The resulting data set of $400,000$ images is then split into $380,000$ and $20,000$ images in a stratified way for training and testing, respectively.
    \end{itemize}
    \item Training methodology\,\textemdash\,For each model, we use the same training approach and parameter values (e.g., learning rate, weight decay, and annealing) as provided in the respective papers. In addition to that, we also experimented with training these models using Adaptive Momentum~\cite{adaptive_momentum} with learning rates of $0.00001$, $0.0001$, and $0.001$, and with a weight decay of $0.0001$. Unless specified otherwise in the respective papers, we train each model for $100$ epochs and provide results for that experiment that achieved the highest overall accuracy on the test sets described above.  
\end{itemize}

\end{document}